\documentclass[journal]{IEEEtran}
\ifCLASSINFOpdf
\else
\fi
\long\def\comment#1{}

\newfont{\bbb}{msbm10 scaled 700}

\newfont{\bb}{msbm10 scaled 1100}

\newcommand{\RR}{\mbox{\bb R}}

\newcommand{\EE}{\mbox{\bb E}}

\newcommand{\mv}{{\bf m}}

\newcommand{\pv}{{\bf p}}
\newcommand{\qv}{{\bf q}}

\newcommand{\wv}{{\bf w}}
\newcommand{\vv}{{\bf v}}
\newcommand{\xv}{{\bf x}}
\newcommand{\yv}{{\bf y}}
\newcommand{\zv}{{\bf z}}
\newcommand{\zerov}{{\bf 0}}

\newcommand{\Am}{{\bf A}}
\newcommand{\Bm}{{\bf B}}

\newcommand{\Id}{{\bf I}}

\newcommand{\Bc}{{\cal B}}

\newcommand{\Ec}{{\cal E}}
\newcommand{\Fc}{{\cal F}}
\newcommand{\Gc}{{\cal G}}
\newcommand{\Hc}{{\cal H}}

\newcommand{\Lc}{{\cal L}}

\newcommand{\Nc}{{\cal N}}
\newcommand{\Oc}{{\cal O}}

\newcommand{\Rc}{{\cal R}}

\newcommand{\Vc}{{\cal V}}
\newcommand{\Xc}{{\cal X}}
\newcommand{\Yc}{{\cal Y}}

\newcommand{\alphav}{\hbox{\boldmath$\alpha$}}

\newcommand{\gammav}{\hbox{\boldmath$\gamma$}}

\newcommand{\lambdav}{\hbox{\boldmath$\lambda$}}

\newcommand{\thetav}{\hbox{\boldmath$\theta$}}

\renewcommand{\arg}{{\hbox{arg}}}

\newcommand{\eqdef}{\stackrel{\Delta}{=}}

\newcommand{\trasp}{{\sf T}}
\newcommand{\transp}{{\sf T}}

\usepackage{array}
\usepackage{multirow}
\usepackage{enumerate}
\usepackage{makecell}
\usepackage{tabularx} 
\usepackage{algorithm,algcompatible}
\usepackage{lipsum}
\usepackage{bbm}
\usepackage{times}
\usepackage[cmex10]{amsmath}
\usepackage{algpseudocode}
\usepackage{epsfig,verbatim}
\usepackage{amssymb} 
\usepackage{graphics,subfigure}
\usepackage{xcolor}
\usepackage{graphicx}
\usepackage{tabularx,booktabs,hhline}
\usepackage{color, colortbl}
\definecolor{LightCyan}{rgb}{0.88,1,1}
\definecolor{lightgray}{gray}{0.95}
\usepackage{multirow}

\newtheorem{theorem}{Theorem}

\newtheorem{lemma}{Lemma}

\newtheorem{corollary}{Corollary}

\newtheorem{remark}{Remark}

\newcommand{\argmin}{\operatornamewithlimits{argmin}}

\usepackage{array}
\usepackage{multirow}
\usepackage{enumerate}
\usepackage{makecell}
\usepackage{tabularx} 
\usepackage{algorithm,algcompatible}
\usepackage{lipsum}
\usepackage{bbm}
\usepackage{times}
\usepackage[cmex10]{amsmath}
\usepackage{algpseudocode}
\usepackage{epsfig,verbatim}
\usepackage{amssymb} 
\usepackage{graphics,subfigure}
\usepackage{xcolor}
\usepackage{graphicx}
\usepackage{tabularx,booktabs,hhline}
\usepackage{color, colortbl}
\definecolor{LightCyan}{rgb}{0.88,1,1}
\definecolor{lightgray}{gray}{0.95}
\usepackage{multirow}

\begin{document}
%
\title{Multiple Kernel-Based Online Federated Learning}

%
%
%

\author{Jeongmin Chae,~\IEEEmembership{Student,~IEEE,}
        and~Songnam Hong,~\IEEEmembership{Member,~IEEE}
\thanks{J. Chae is with the Department of Electrical Engineering, University of Southern California, CA, 90089, USA (e-mail: chaej@usc.edu)}
\thanks{S. Hong is with the Department of Electronic Engineering, Hanyang University, Seoul, 04763, Korea (e-mail: snhong@hanyang.ac.kr)} 

}

\maketitle

\begin{abstract}
Online federated learning (OFL) becomes an emerging learning framework, in which edge nodes perform online learning with continuous streaming local data and a server constructs a global model from the aggregated local models. Online multiple kernel learning (OMKL), using a preselected set of $P$ kernels, can be a good candidate for OFL framework as it has provided an outstanding performance with a low-complexity and scalability. Yet, an naive extension of OMKL into OFL framework suffers from a heavy communication overhead that grows linearly with $P$. In this paper, we propose a novel multiple kernel-based OFL (MK-OFL) as a non-trivial extension of OMKL, which yields the same performance of the naive extension with $1/P$ communication overhead reduction. We theoretically prove that MK-OFL achieves the optimal sublinear regret bound when compared with the best function in hindsight. Finally, we provide the numerical tests of our approach on real-world datasets, which suggests practicality.
\end{abstract}

%
%
%
\section{Introduction}
\label{intro}


Federated learning is an emerging distributed learning framework, in which distributed nodes (e.g., mobile phones, wearable devices, etc.) train a model collaboratively under the coordination of a central server without centralizing the training data \cite{hard2018federated, kairouz2019advances}. To be specific, federated learning optimizes a global model by repeating the two operations: i) local model optimizations at edge nodes; ii) global model update (e.g., model averaging) at the server \cite{yang2019federated}. This approach has received tremendous attention due to the myriad of applications: activities of mobile phone users, predicting a low blood sugar, heart attack risk from wearable devices, or detecting burglaries within smart homes \cite{anguita2013public, pantelopoulos2009survey, rashidi2009keeping}.

In many real-world applications, machine learning tasks are expected to be operated in an online fashion. For example, online learning is required when data is generated as a function of time (i.e., time-series predictions) \cite{richard2008online,shen2012stock} and when the large number of data makes it impossible to carry out data analytic in batch form \cite{kivinen2004online}. This challenging problem has been successfully addressed via online multiple kernel learning (OMKL) \cite{ shawe2004kernel, shen2019random,hong2020active}. OMKL learns a sequence of functions (or models) which predicts the label of a newly incoming data. In particular, it seeks the optimal combination of a pool of $P$ single kernel functions in an online fashion. With the use of multiple kernels, OMKL can provide a superior accuracy and enjoy a great flexibility compared with single-kernel online learning \cite{shen2019random,hong2020active}. In spite of these merits, OMKL is restricted to a centralized online learning and an extension to a federated learning is still open problem.


In this paper, we consider an online federated learning (OFL) problem. The objective of OFL is to learn a sequence of global models using continuous streaming data across a large number of distributed nodes. At every time, specifically, OFL optimizes a global model by repeating two operations: i) edge nodes updates local models using their own incoming data; ii) the server constructs an improved global model by averaging the them. The construction of a single kernel-based OFL (named SK-OFL) is rather straightforward, where the parameter of a kernel function (defined by a $2D$-dimensional vector) is simply exchanged between every node and the server. As in the centralized counterpart \cite{shen2019random,hong2020active}, the use of multiple kernels (i.e., a preselected set of $P$ kernels) is necessary to enhance a learning accuracy. However, a naive extension of SK-OFL into a multiple kernel setting suffers from a heavy communication overhead that increases linearly with $P$. This is because the parameters of all $P$ kernels should be exchanged. In this paper, we propose a novel multiple-kernel OFL (named MK-OFL) which can achieve the same performance of the naive extension and outperforms SK-OFL considerably with the same communication overhead with SK-OFL. To this end, our major contributions are summarized as follows.
\begin{itemize}

    \item In the proposed MK-OFL, every node selects one kernel index out of $P$ kernels randomly according to its own probability mass function (PMF). Then, it sends one local model corresponding to the selected index, thus having the same communication overhead with SK-OFL. With high probability, this randomized algorithm chooses the best kernel only after a certain convergence time, thus being able to yield an attractive performance.
    
    \item Leveraging the martingale argument, we theoretically prove that MK-OFL achieves the optimal sublinear regret bound $\Oc(\sqrt{T})$ when compared wit the best kernel function in hindsight. This implies that MK-OFL achieves the same accuracy of the centralized OMKL asymptotically by preserving an edge-node privacy.
    
    \item Via numerical tests on real-world datasets, we demonstrate the effectiveness of the proposed MK-OFL in various online learning tasks such as online regressions and time-series predictions. Remarkably, it is shown that MK-OFL significantly outperforms SK-OFL and almost achieves the performance of the best kernel function in hindsight.
\end{itemize}

%
%
\section{Preliminaries}\label{sec:preliminaries}

We briefly review the kernel-based online learning framework (named OMKL) in \cite{hong2020active}, and describe the problem setting of an online federated learning (OFL). To simplify the notation, we let $[N]=\{1,2,...,N\}$ for some positive integer $N$. Also, $t$, $k$, $p$ represent the indices of time, node, and kernel, respectively.

\subsection{Overview of a kernel-based online learning}

The main objective of an online learning is to learn a sequence of functions $\{\hat{f}_{t}: t\in[T]\}$ such that the cumulative regret is minimized:
\begin{equation}
    \Rc_T=\sum_{t=1}^{T} \Lc(\hat{f}_{t}(\xv_t),y_t) - \sum_{t=1}^{T}\Lc(f^{\star}(\xv_t),y_t), \label{eq:regret_loss}
\end{equation} where $\Lc(\cdot,\cdot)$ and $f^{\star}(\cdot)$ denote a cost function and the best function in hindsight, respectively. Specifically, at every time $t-1$, $\hat{f}_{t}(\cdot)$ is estimated from the received data $\{(\xv_\tau, y_\tau): \tau \in [t-1]\}$. The above challenging problem has been efficiently solved via a kernel-based online learning \cite{shen2019random,hong2020active}. In this approach, a function space for an optimization is restricted to a reproducing Hilbert kernel space (RKHS) $\Hc_p$, defined as 
$\Hc_p \eqdef \left\{f: f(\xv)=\sum_{t}\alpha_{t}\kappa_p(\xv,\xv_{t})\right\}$,
where $\kappa_p(\xv,\xv_{t}): \Xc \times \Xc \rightarrow \Yc$ is a symmetric positive semidefinite basis function (called kernel) \cite{wainwright2019high}. Moreover, the representer theorem \cite{wahba1990spline} and random-feature approximation \cite{rahimi2007random} shows that each kernel function in an optimal solution of (\ref{eq:regret_loss}) can be well-approximated with a model parameter $\hat{\wv}_t^p$, i.e., 
\begin{equation}
    \hat{f}_{t}(\xv) = \left(\hat{\wv}_{t}^{p}\right)^{\trasp}\zv_{p}(\xv),\label{eq:single_f}
\end{equation}where $\zv_p(\xv)$ denotes a non-linear function defined by a kernel $\kappa_{p}$, i.e.,
\begin{align}
    &\zv_p(\xv)=\label{eq:z}\\
    &\frac{1}{\sqrt{D}}[\sin\vv_{1}^{\trasp}\xv, \dots,\sin\vv_{D}^{\trasp}\xv,\cos\vv_{1}^{\trasp}\xv,\dots,\cos\vv_{D}^{\trasp}\xv]^{\trasp},\nonumber
\end{align} and where $\{\vv_i: i\in [D]\}$ denotes an independent and identically distributed (i.i.d.) samples from the Fourier transform of a given kernel function $\kappa_p(\cdot,\cdot)$. Note that the hyper-parameter $D$ can be chosen independently from the size of data $T$. The accuracy of the above kernel-based online learning fully relies on a preselected kernel $\kappa_p$, which is chosen manually either by a task-specific priori knowledge or by some intensive cross-validation process. Online multiple kernel learning (OMKL), using a predetermined set of $P$ kernels (called a kernel dictionary), is more powerful \cite{shen2019random, hong2020active}. Namely, OMKL seeks a sequence of best kernel functions, each of which has the following form:
 \begin{align}\label{eq:mkl_app}
    \hat{f}_t(\xv)=\sum_{p=1}^{P}\hat{q}_{t}^{p}\left(\hat{\wv}_{t}^{p}\right)^{\trasp}\zv_{p}(\xv) \in \bar{\Hc}, 
\end{align} where  $\bar{\Hc}\eqdef \Hc_1 + \Hc_2 + \cdots + \Hc_P$ is again RKHS \cite{wainwright2019high} and $\hat{q}_t^p\in[0,1]$ denotes the combination weight (or reliability) of the associated kernel function $\hat{f}_t^p(\xv)=\left(\hat{\wv}_{t}^{p}\right)^{\trasp}\zv_{p}(\xv)$. 
We remark that the optimization of OMKL can be solved using the powerful tool boxes from online convex optimization and online learning techniques developed over vector spaces.

\begin{figure*}[!t]
\centerline{\includegraphics[width=15cm]{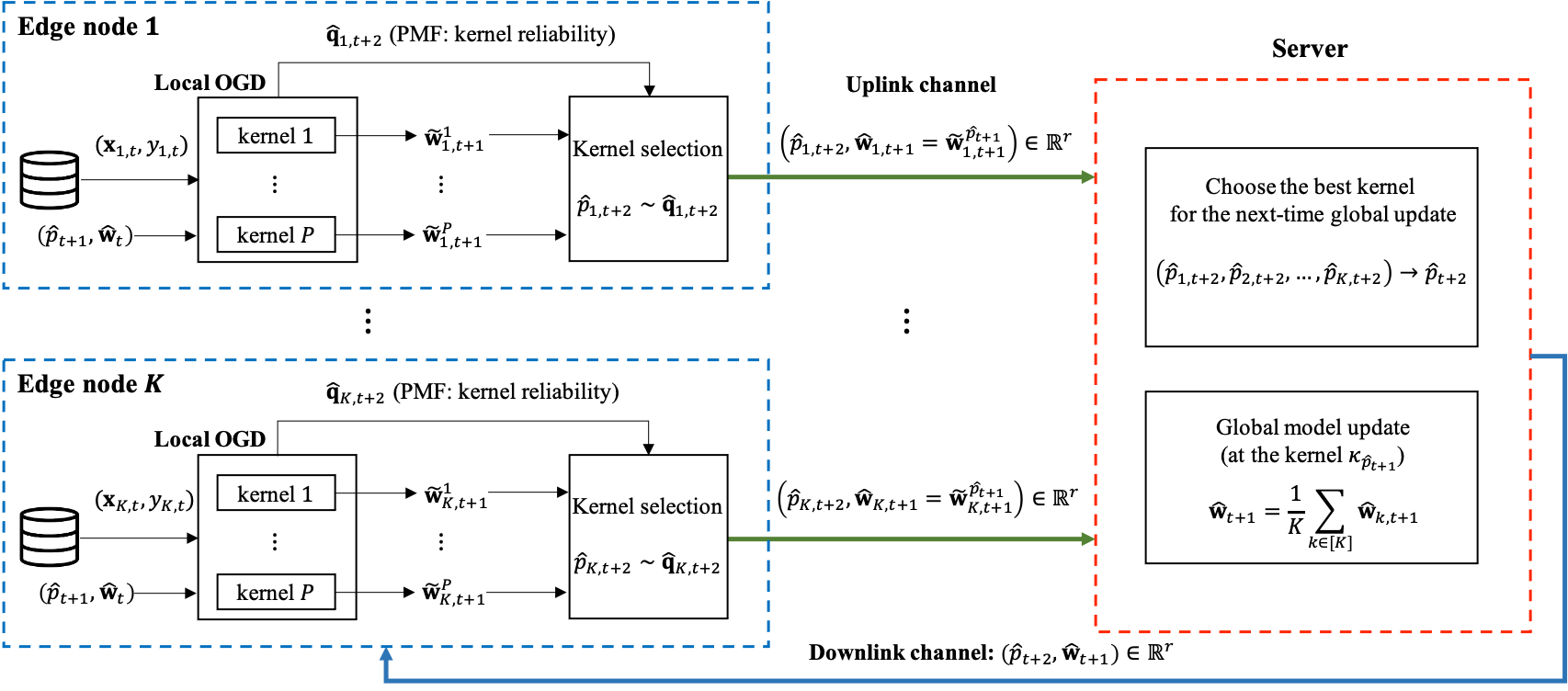}}
\caption{Description of the proposed online federated learning with multiple kernels at time $t$, in which $\hat{\wv}_{k,t+1}$ and $\hat{\wv}_{t+1}$ denote the local model of the node $k$ and the global model, respectively. The estimated label $\hat{y}_{k,t}$ at each node $k$ is determined as $\hat{y}_{k,t}=\hat{\wv}_{t}^{\trasp}\zv_{\hat{p}_t}(\xv_{k,t})$.}
\label{fig:MK-OFL}
\end{figure*}

%
%
\subsection{Problem Formulation}\label{subsec:PF}

We define the problem setting of an online federated learning (OFL). The objective of OFL is to learn a sequence of functions (i.e., global models) from sequentially incoming data across a large number of distributed nodes. In detail, at each time $t$, the server distributes the current global model $\hat{\wv}_{t} \in \RR^r$ (the parameter of a learned function) to the distributed $K$ nodes, where $r$ denotes a communication overhead. Every nodes $k$ updates the local model independently from the current global model and incoming data $(\xv_{k,t},y_{k,t})$, where $\xv_{k,t}\in \Xc \subseteq \RR^d$ and $y_{k,t}\in\Yc\subseteq \RR$ represent the feature and the label, respectively. Hereinafter, the corresponding local model are denoted as $\hat{\wv}_{k,t+1}$ for $k\in [K]$.  Then, the $K$ nodes send it back to the server, from which the server constructs an improved global model as
\begin{equation}
    \hat{\wv}_{t+1} = h\left(\hat{\wv}_{1,t+1},...,\hat{\wv}_{K,t+1}\right),
\end{equation} for some function $h: \RR^{r}\times\cdots\times\RR^{r}\rightarrow \RR^{r}$. In this paper, it is assumed that $h$ is the popular {\em averaging function} (FedAvg) \cite{konevcny2016federated}, i.e., $\hat{\wv}_{t+1}=\frac{1}{K}\sum_{k=1}^{K} \hat{\wv}_{k,t+1}$.
These two-step procedures will be repeated.

%
%
\section{Methods}\label{sec:methods}

We first present a single kernel-based OFL (SK-OFL) and then extend it into a multiple kernel setting. The resulting method is named MK-OFL. To meet the communication overhead, in both methods, the messages exchanged between a server and edge nodes are constructed as $r$-dimensional real-valued vectors (see Fig.~\ref{fig:MK-OFL}). The detailed algorithms of SK-OFL and MK-OFL are described in Section~\ref{subsec:SK-OFL} and Section~\ref{subsec:MK-OFL}, respectively.

\subsection{The Proposed SK-OFL}\label{subsec:SK-OFL}

The objective of SK-OFL is to seek a sequence of functions $\hat{f}_{1}(\xv),...,\hat{f}_{T}(\xv)$ under the OFL framework in Section~\ref{subsec:PF}. Following the structure of a kernel-based online learning \cite{hong2020active}, each function is fully defined with a single parameter $\hat{\wv}_{t} \in \RR^{2D}$, i.e.,
\begin{equation}
    \hat{f}_{t}(\xv) = \hat{\wv}_{t}^{\trasp}\zv_{p}(\xv),
\end{equation}  where $\zv_{p}(\xv)$ is defined in \eqref{eq:z} and a preselected kernel $\kappa_p$ is assumed. In SK-OFL, $\hat{\wv}_{t}$ is considered as a global model and by setting $D=\lfloor r/2 \rfloor$, the communication overhead is met. At every time $t$, SK-OFL performs a distributed model training iteratively with the two steps: i) local model update at the edge nodes; ii) global model update at the server. Focusing on time $t$, we provide the specific procedures of these steps below.

{\em (i) Local model update:} Given the current global model $\hat{\wv}_{t}$ and an incoming data $(\xv_{k,t},y_{k,t})$, each node $k$ updates its local model $\hat{\wv}_{k,t+1}$ via online gradient descent (OGD) \cite{hazan2016introduction}:
\begin{equation}
    \hat{\wv}_{k,t+1} = \hat{\wv}_{t} - \eta_{\ell} \nabla\Lc\left(\hat{\wv}_{t}^{\trasp}\zv_{p}(\xv_{k,t}),y_{k,t} \right),
\end{equation} with step size $\eta_l>0$, where $\nabla\Lc\left(\hat{\wv}_{t}^{\trasp}\zv_{p}(\xv_{k,t}),y_{k,t} \right)$ stands for the gradient at the point $\hat{\wv}_{t}$. Then, it sends the updated local model $\hat{\wv}_{k,t+1}$ to the server.

{\em (ii) Global model update:} Given the received local models 
$\{\hat{\wv}_{k,t+1}: k\in[K]\}$, the server updates the global model by averaging them: 
\begin{equation}
    \hat{\wv}_{t+1} = \frac{1}{K} \sum_{k=1}^{K} \hat{\wv}_{k,t+1}.
\end{equation} Then, it distributes the updated global model $\hat{\wv}_{t+1}\in\RR^{r}$ to the $K$ nodes.

\begin{remark} The proposed SK-OFL can be immediately extended into a multiple kernel setting, in which every node transmits the $P$ local models and the corresponding weights as in \eqref{eq:mkl_app} to the server, and vice versa. This naive extension suffers from an expensive communication overhead as it linearly increases with the size of a kernel dictionary $P$ (i.e., equals to $P\times r$). In the following section, we will present a novel MK-OFL as a non-trivial extension of SK-OFL into a multiple kernel setting. It is noticeable that MK-OFL can achieve the performance of the naive extension and significantly improve the performance of SK-OFL while having the communication overhead $r$ irrespective of $P$.
\end{remark}

%
%
\subsection{The Proposed MK-OFL}\label{subsec:MK-OFL}

In the proposed MK-OFL, the central server broadcasts the global information $(\hat{p}_{t+1}, \hat{\wv}_{t}) \in [P] \times \RR^{2D}$ at time $t$ over $t=1,...,T$, where the initial values are assigned as 
$\hat{\wv}_{1} = \zerov$ and $\hat{p}_{2}=1$. Regarding the communication overhead, compared with SK-OFL a kernel index is additionally transmitted.
Taking this into account, $(\hat{p}_{t+1},\hat{\wv}_{t})$ is considered as a global model and by setting $D=\lfloor r/2 \rfloor -1$, the communication overhead is satisfied. Then, a common learned (or estimated) function at the current time $t$ is fully determined by the parameters $\hat{p}_{t}$ (obtained at time $t-1$) and $\hat{\wv}_{t}$, i.e., 
\begin{equation}
    \hat{f}_{t}(\xv) = \hat{\wv}_{t}^{\trasp}\zv_{\hat{p}_{t}}(\xv).
\end{equation}  The key idea of MK-OFL is following: it is allowed to choose a different kernel at every time, while as $t$ grows, a selected kernel can converge to the best kernel in hindsight. Furthermore, we verified that the convergence speed is quite fast. In contrast, SK-OFL should choose one kernel before receiving data and thus, the predetermined kernel is unlikely to be the best one. For these reasons, MK-OFL can significantly outperform SK-OFL without sacrificing the communication overhead and by enjoying the use of multiple kernels. Focusing on time $t$, we provide the detailed algorithm of MK-OFL below (see Fig.~\ref{fig:MK-OFL}). 

{\em (i) Local model update:} At time $t$, each node $k$ has its own local information as $\{\tilde{\wv}_{k,t}^{p}: p\in[P]\}$ and receives the global model $(\hat{p}_{t+1},\hat{\wv}_{t})$ from the server. Using the $\hat{p}_{t}$ (observed at time $t-1$) and $\hat{\wv}_{t}$, it first updates the parameters of the $P$ local kernel functions:
\begin{equation}\label{eq:w_update}
    \hat{\wv}_{k,t}^p = 
    \begin{cases}
    \tilde{\wv}_{k,t}^p, \mbox{ if } p=\hat{p}_{t}\\
    \hat{\wv}_{t}, \mbox{ if } p\neq\hat{p}_{t},
    \end{cases}
\end{equation} for $\forall p \in [P]$. Then, every node $k$ updates the local information via OGD:
\begin{equation}
    \tilde{\wv}_{k,t+1}^{p}=\hat{\wv}_{k,t}^{p} -\eta_{\ell} \nabla\Lc\left((\hat{\wv}_{k,t}^{p})^{\trasp}\zv_{p}(\xv_{k,t}),y_{k,t} \right),\label{eq:localupdate}
\end{equation} for $\forall p \in [P]$, where $\eta_{\ell}>0$ is the step size. Given the received parameter $\hat{p}_{t+1}$, set
\begin{equation}
    \hat{\wv}_{k,t+1}=\tilde{\wv}_{k,t+1}^{\hat{p}_{t+1}},
\end{equation} which will be transmitted to the server as the updated local model. 

We next explain how each node $k$ selects the candidate of the best kernel at time $t+2$ (denoted by $\hat{p}_{k,t+2}$).
The weights for combining the $P$ kernel functions are determined on the basis of the past local losses, i.e., 
\begin{equation}\label{eq:prob}
    \hat{\qv}_{k,t+2}(p) = \frac{\hat{\mv}_{k,t+2}(p)}{\sum_{p=1}^P \hat{\mv}_{k,t+2}(p)}, \;\; \forall p \in [P],
\end{equation} where the initial values $\hat{\mv}_{k,2}(p)=1, \forall p \in [P]$ and for a learning rate $\eta_g>0$, 
\begin{align}
    &\hat{\mv}_{k,t+2}(p)= \hat{\mv}_{k,t+1}(p)\nonumber\\
    &\;\;\;\;\; \times \exp\left(-\eta_g K\Lc\left((\hat{\wv}_{k,t}^{p})^{\trasp}\zv_{p}(\xv_{k,t}), y_{k,t}\right)\right).\label{eq:local_loss}
\end{align} In the context of online learning, the above weight update is known as {\em exponential strategy} (or Hedge algorithm) \cite{bubeck2011introduction}. From \eqref{eq:prob}, the following probability mass function (PMF) is defined
\begin{equation}\label{eq:pmf}
\hat{\qv}_{k,t+2} = (\hat{\qv}_{k,t+2}(1),...,\hat{\qv}_{k,t+2}(P)).
\end{equation} Then, it selects one kernel index $\hat{p}_{k,t+2} \in [P]$ according to the PMF $\hat{\qv}_{k,t+2}$ and transmits the updated local model  $(\hat{p}_{k,t+2},\hat{\wv}_{k,t+1}=\tilde{\wv}_{k,t+1}^{\hat{p}_{t}})$ to the server. 

{\em (ii) Global model update:} The server receives the updated local models $\{(\hat{p}_{k,t+2},\hat{\wv}_{k,t+1}): k\in[K]\}$ from the $K$ nodes. First, the global parameter is updated by averaging the local parameters such as
\begin{equation}
    \hat{\wv}_{t+1} = \frac{1}{K}\sum_{k=1}^{K} \hat{\wv}_{k,t+1}.
\end{equation} Also, each kernel $k$ chooses one kernel index  $\hat{p}_{t+2}$ from $\{\hat{p}_{k,t+2}: k\in[K]\}$ randomly according to the PMF $\hat{\alphav}_{t+2}=(\hat{\alpha}_{1,t+2},...,\hat{\alpha}_{K,t+2})$, where
\begin{align}
\alpha_{k,t+2} &= \frac{|\{k\in[K]: \hat{p}_{k,t+1}=p\}|^{K-1}}{\sum_{p=1}^{P}{|\{k\in[K]: \hat{p}_{k,t+1}=p\}|^K}}. \label{eq:weight}
\end{align} Naturally, this PMF construction aims at assigning a higher probability to a more frequent index from $\{\hat{p}_{k,t+2}: k\in[K]\}$. The resulting index $\hat{p}_{t+2}$ represents the kernel index to be globally updated at the next time slot $t+2$. The server distributes the updated global model $(\hat{p}_{t+2}, \hat{\wv}_{t+1})$ to the $K$ nodes.

\section{Regret Analysis}\label{sec:anal}

We analyze the cumulative regret of the proposed MK-OFL. As usually considered for the analysis of online convex optimizations and online learning frameworks \cite{bubeck2011introduction,hazan2016introduction,shen2019random,hong2020active}, the following conditions are also assumed.

\vspace{0.1cm}
{\bf Assumption 1.} {\it For any fixed $\zv_{p}(\xv_t)$ and $y_t$, the loss function $\Lc(\wv^{\trasp}\zv_{p}(\xv_{t}), y_t)$ is convex with respect to $\wv$, and is bounded as $\Lc(\wv^{\trasp}\zv_{p}(\xv_t),y_t) \in [0,1]$.} 

\vspace{0.1cm}
{\bf Assumption 2.}
{\it For any kernel $\kappa_p$, $\wv_{k,t}^p$ belongs to a bounded set $\Bc$, i.e., $\|\wv_{k,t}^p\| \leq C$ for any $t \in [T]$.}

\vspace{0.1cm}
{\bf Assumption 3.}
{\it The loss function is $L$-Lipschitz continuous, i.e., $\|\nabla \Lc(\wv^{\trasp}\zv_{p}(\xv_t),y_t)\| \leq L$.}

Let $f_{p}^{\star}(\xv)=(\wv_{p}^{\star})^{\trasp}\zv_p(\xv)$ denote the best function in the kernel $\kappa_p$ with respect to the distributed data 
$\{(\xv_{k,t},y_{k,t}): k\in[K], t\in[T]\}$:
\begin{equation}\label{eq:optimal}
    \wv_{p}^{\star} \eqdef \arg\min_{\wv\in\Bc} \sum_{t=1}^{T}\sum_{k=1}^{K} \Lc\left(\wv^{\trasp}\zv_{p}(\xv_{k,t}), y_{k,t}\right).
\end{equation}

Before stating our main results, we will give some intuitions behind the proposed MK-OFL. It is a {\em randomized} algorithm due to the randomness of the best kernel selection at every time $t$ (denoted by $\hat{p}_t$ in our algorithm). By integrating randomness at the distributed edges and the server, it can be understood that $\hat{p}_{t}$ is chosen from $[P]$ randomly according to the following PMF:
\begin{equation}
    \hat{\qv}_{t} =\sum_{k=1}^{K}\alpha_{k,t}\hat{\qv}_{k, t},
\end{equation} where $\alpha_{k,t}$, defined in \eqref{eq:weight}, can capture the reliabilities of the $K$ nodes at the current time $t$. To evaluate the accuracy of $\hat{\qv}_{t}$, we introduce the corresponding network-wise (or centralized) PMF $\bar{\qv}_{t}$, which is determined on the basis of the entire losses of the $K$ nodes:
\begin{equation}\label{eq:overallPMF}
    \bar{\qv}_{t}(p) = \frac{\bar{\mv}_{t}(p)}{\sum_{p=1}^P \bar{\mv}_{t}(p)},\;\; \forall p \in [P],
\end{equation} with the initial values $\bar{\mv}_{1}(p)=1, \forall p \in [P]$, where 
\begin{equation}
    \bar{\mv}_{t}(p) = \bar{\mv}_{t-1}(p)\times L_{t-1}^p,
\end{equation} and for some parameter $\eta_g>0$
\begin{equation*}
    L_{t-1}^p =\prod_{k=1}^{K} \exp\left(-\eta_g\Lc\left((\hat{\wv}_{k,t-1}^{p})^{\trasp}\zv_{p}(\xv_{k,t-1}), y_{k,t-1}\right) \right).
\end{equation*} In general, the data available locally fail to estimate the overall distribution $\bar{\qv}_t$. Obviously MK-OFL can approach the performance of the centralized OMKL \cite{hong2020active} (i.e., the performance limit), provided that our PMF $\hat{\qv}_{t}$, which is constructed in a decentralized way, is sufficiently close to $\bar{\qv}_{t}$.  In fact, an exact gap between them relies on the characteristics (e.g., heterogeneity) of real datasets. Nevertheless, in our theoretical analysis, it is proved that regardless of datasets, the above gap only leads to a bounded performance loss, which is less than $\Oc(\sqrt{T})$. Namely, the loss is small enough to guarantee the optimal sublinear regret $\Oc(\sqrt{T})$ of the proposed MK-OFL. Beyond the asymptotic analysis, our experiments in Fig. 2 demonstrate that the impact of the above gap is negligible.

\subsection{Main results}

Note that in the proposed MK-OFL, the server broadcasts the current estimate $\hat{wv}_{t}$ at time $t$ to the $K$ nodes and each node $k$ estimates the label of a newly incoming data $\xv_{k,t}$ as $\hat{y}_{k,t}=\hat{\wv}_{t}^{\trasp}\zv_{\hat{p}_t}(\xv_{k,t})$. Also, $\wv_{p}^{\star}$ denotes the optimal parameter defined in \eqref{eq:optimal}. Under the Assumption 1 - Assumption 3, we state the main result of this section:

\begin{theorem}\label{thm1} The proposed MK-OFL with 
$\eta_{\ell}=\eta_g = \Oc(1/\sqrt{T})$ guarantees the sublinear regret $\Oc(\sqrt{T})$ with high probability:
\begin{align*}
    \Rc_{T} &=\sum_{t=1}^{T}\sum_{k=1}^{K}\Lc\left(\hat{\wv}_{t}^{\trasp}\zv_{\hat{p}_t}(\xv_{k,t}), y_{k,t}\right)\nonumber\\
    & - \min_{1\leq p \leq P}\sum_{t=1}^{T}\sum_{k=1}^{K} \Lc\Big((\wv_{p}^{\star})^{\trasp}\zv_{p}(\xv_{k,t}), y_{k,t}\Big)\leq \Oc(\sqrt{T}).
\end{align*} 
\end{theorem}\vspace{0.2cm} 

From the fact that SK-OFL is a deterministic algorithm with a preselected kernel $\kappa_p$, we can immediately get:
%
%
\begin{corollary}\label{cor1} For a predetermined kernel $\kappa_p$, SK-OFL with $\eta_{\ell}=\eta_g = \Oc(1/\sqrt{T})$ guarantees the sublinear regret as

\begin{align*}
    \Rc_{T} &=\sum_{t=1}^{T}\sum_{k=1}^{K}\Lc\left(\hat{\wv}_{t}^{\trasp}\zv_{p}(\xv_{k,t}), y_{k,t}\right)\nonumber\\
    &\;\;\;\;\;\; - \sum_{t=1}^{T}\sum_{k=1}^{K}\Lc\Big((\wv_{p}^{\star})^{\trasp}\zv_{p}(\xv_{k,t}), y_{k,t}\Big)\leq \Oc(\sqrt{T}).
\end{align*}
\end{corollary}

Theorem~\ref{thm1} and Corollary~\ref{cor1} reveal that MK-OFL and SK-OFL can achieve the optimal sublinear regret bounds when compared with the respective best functions in hindsight. However, it is noticeable that the best functions as to MK-OFL and SK-OFL are from  $\bar{\Hc}= \Hc_1 + \cdots + \Hc_P$ and $\Hc_{p} \subset \bar{\Hc}$ for a preselected $p \in [P]$, respectively.
Consequently, the former tends to be much better than the latter. For this reason, MK-OFL can considerably outperform SK-OFL, which will be demonstrated in Section~\ref{sec:exp} via online learning tasks with real-world datasets.

\subsection{Proof of Theorem 1}
To capture the randomness of kernel selections in MK-OFL, we introduce a random variable $I_t$ which indicates a selected kernel index at time $t$ in the server. As in the description of MK-OFL, the realization of $I_t$ is denoted as $\hat{p}_{t}$. The parameters of $P$ kernel functions and the local PMFs are also random variables, which are determined as a function of random variables $I_1,...,I_{t-1}$. Without loss of generality, their realizations at time $t$ are denoted as 
$\{\hat{q}_{k,t}(p):p\in[P], k\in[K]\}$ and $\{\hat{\wv}_{k,t}^p: p\in [P], k\in[K]\}$, respectively. Also, the realizations of the PMFs are denoted as $\hat{\qv}_{t}(p)$ and $\bar{\qv}_{t}(p)$. With these notations, we first provide key lemmas for this proof. Lemma~\ref{lem1} are immediately obtained following the proof of OGD convergence \cite{hazan2016introduction} with the Assumption 2 and Assumption 3. The proofs of Lemma~\ref{lem2} and Lemma~\ref{lem3} are provided in the appendices.

\vspace{0.2cm} 
\begin{lemma}\label{lem1} For any $p \in [P]$ and step size $\eta_l>0$, the following upper bound is hold:
\begin{align*}
    &\sum_{t=1}^{T}\sum_{k=1}^K\Lc\left(\left(\hat{\wv}_{k,t}^{p}\right)^{\trasp}\zv_{p}(\xv_{k,t}),y_{k,t}\right) \nonumber\\
    &\;\;\;\;\;\;\;\;\;\;\;\;\;\;\;\;\;\;\;\;\;\;\;\; - \sum_{t=1}^{T}\sum_{k=1}^K \Lc\Big((\wv_{p}^{\star})^{\trasp}\zv_{p}(\xv_{k,t}),y_{k,t}\Big)\nonumber\\
    &\leq \frac{K C^2}{2\eta_l} + \frac{3\eta_l K L^2 T}{2}.
\end{align*}
\end{lemma}
{\bf Proof:} The proof is provided in Appendix~\ref{app:proof1}.
\vspace{0.2cm}


\begin{lemma}\label{lem2} For any fixed parameter $\eta_g >0$, we have:
\begin{align*}
    &\sum_{t=1}^{T}\sum_{k=1}^{K}\Lc\left(\sum_{p=1}^{P}\hat{q}_{t}(p) \left(\hat{\wv}_{k,t}^p\right)^{\trasp}\zv_{p}(\xv_{k,t}),y_{k,t}\right)\nonumber\\
    &\;\;\;\;\;\;\;\;\;\;\;\;\;\;\;\;\;\;\; - \min_{1\leq p \leq P} \sum_{t=1}^T \sum_{k=1}^{K} \Lc\left( \left(\hat{\wv}_{k,t}^p\right)^{\trasp}\zv_{p}(\xv_{k,t}),y_{j,t}\right)\\
    &\leq \frac{1}{\eta_g} \log{P} + \frac{9 \eta_g  K T}{8}.
\end{align*}
\end{lemma}
{\bf Proof:} The proof is provided in Appendix~\ref{app:proof2}.

It is remarkable that Lemma 1 and Lemma 2 hold for any realization of our randomized algorithm. Setting $\eta_l = \Oc(1\sqrt{T})$, the upper bound of Lemma~\ref{lem1} is bounded by $\Oc(\sqrt{T})$.  Also, setting $\eta_g = \Oc(1/\sqrt{T})$, the upper bound of Lemma~\ref{lem2} is bounded by $\Oc(\sqrt{T})$. We next show that our randomized algorithm, choosing one kernel at every time instead of using the combination of all $P$ kernels, only leads to a bounded loss $\Oc(\sqrt{T})$.

\vspace{0.2cm} 
\begin{lemma}\label{lem3} For some $\delta>0$, the following bound holds with at least probability $1-\delta$:
\begin{align*}
&\sum_{t=1}^{T}\sum_{k=1}^{K}\Lc\Big(W_t^{\trasp}\zv_{I_t}(\xv_{k,t}),y_{k,t}\Big) \\
&\;\;\;\;\;\;\;\;\;\;\;\; -\sum_{t=1}^{T}\sum_{k=1}^{K}\Lc\left(\sum_{p=1}^{P}\hat{q}_{t}(p)\left(\hat{\wv}_{k,t}^p\right)^{\trasp}\zv_{p}(\xv_{k,t}),y_{k,t}\right)\nonumber\\
&\leq K\sqrt{\frac{\log(\delta^{-1})}{2}T}.
\end{align*}
\end{lemma}
{\bf Proof:} The proof is provided in Appendix~\ref{app:proof3}.

Finally, from Lemma~\ref{lem1}, Lemma~\ref{lem2} and Lemma~\ref{lem3} with the parameters $\eta_l=\eta_g=\Oc(1/\sqrt{T})$, the proof is completed.


%
%
\section{Experimental Results}\label{sec:exp}

We evaluate the accuracy performances of the proposed MK-OFL for various online learning tasks such as online regressions and time-series predictions. The OFL system with $K=20$ distributed nodes and communication overhead $r=100$ is considered, and a regularized least-square loss function is assumed, i.e., 
\begin{equation}
    \Lc(\wv^{\trasp}\zv_{p}(\xv), y) = \left[\wv^{\trasp}\zv_{p}(\xv) - y \right]^2 + \lambda\|\wv\|^2.
\end{equation} Taking the randomness of our algorithm into account, the averaged performances over 50 trials are evaluated.  The learning accuracy at time $t$ is measured by the mean-square-error (MSE) as 
\begin{equation}
    {\rm MSE}(t)=\frac{1}{tK}\sum_{\tau=1}^{t}\sum_{k=1}^{K}(\hat{y}_{k,\tau} - y_{k,\tau})^2,
\end{equation} where $\hat{y}_{k,\tau}$ and $y_{k,\tau}$ denote a predicted label and a true label, respectively. In these experiments, the following hyper-parameters for both SK-OFL and MK-OFL are used:
\begin{equation}
    \eta_{l}=1/\sqrt{t},\; \eta_{g}=\log{P}/\sqrt{t}, \mbox{ and } \lambda=0.01.\label{eq:hyper}
\end{equation} These choices are based on anytime strategy \cite{bubeck2011introduction}, which is widely used in the context of online learning when $T$ is unknown (e.g., continuous streaming data). We remark that this anytime strategy also achieves the same asymptotic performance in Theorem~\ref{thm1}, which can be proved by incorporating the proof-technique of anytime strategy in \cite{bubeck2011introduction} into our proof in Section~\ref{sec:methods}. In non-asymptotic cases, whereas, the performance MK-OFL can be further enhanced by carefully optimizing the hyper-parameters $\eta_{l}$ and $\eta_g$. As noticed in \cite{hong2020active}, such hyper-parameter optimization is still an open problem even in the centralized setting. Instead, in our experiments, one pair of the parameters in \eqref{eq:hyper} are used for all test datasets. We build the kernel dictionary with $11$ Gaussian kernels (i.e., $P=11$), each of which is defined with the basis kernels
\begin{equation}
    \kappa_{p}(\xv,\xv')=\exp\left(-\frac{\|\xv-\xv'\|^2}{2\sigma_p^2}\right),
\end{equation} with the parameters (or bandwidths) $\sigma_{p}^2 = 10^{p-6}$, $p=1,2,...,11$. Finally,  the real-world datasets for our experiments in online regressions and time-series predictions are described in Section~\ref{subsec:OR} and~\ref{subsec:TP}, respectively.


\begin{figure}[!t]
\centerline{\includegraphics[width=9cm]{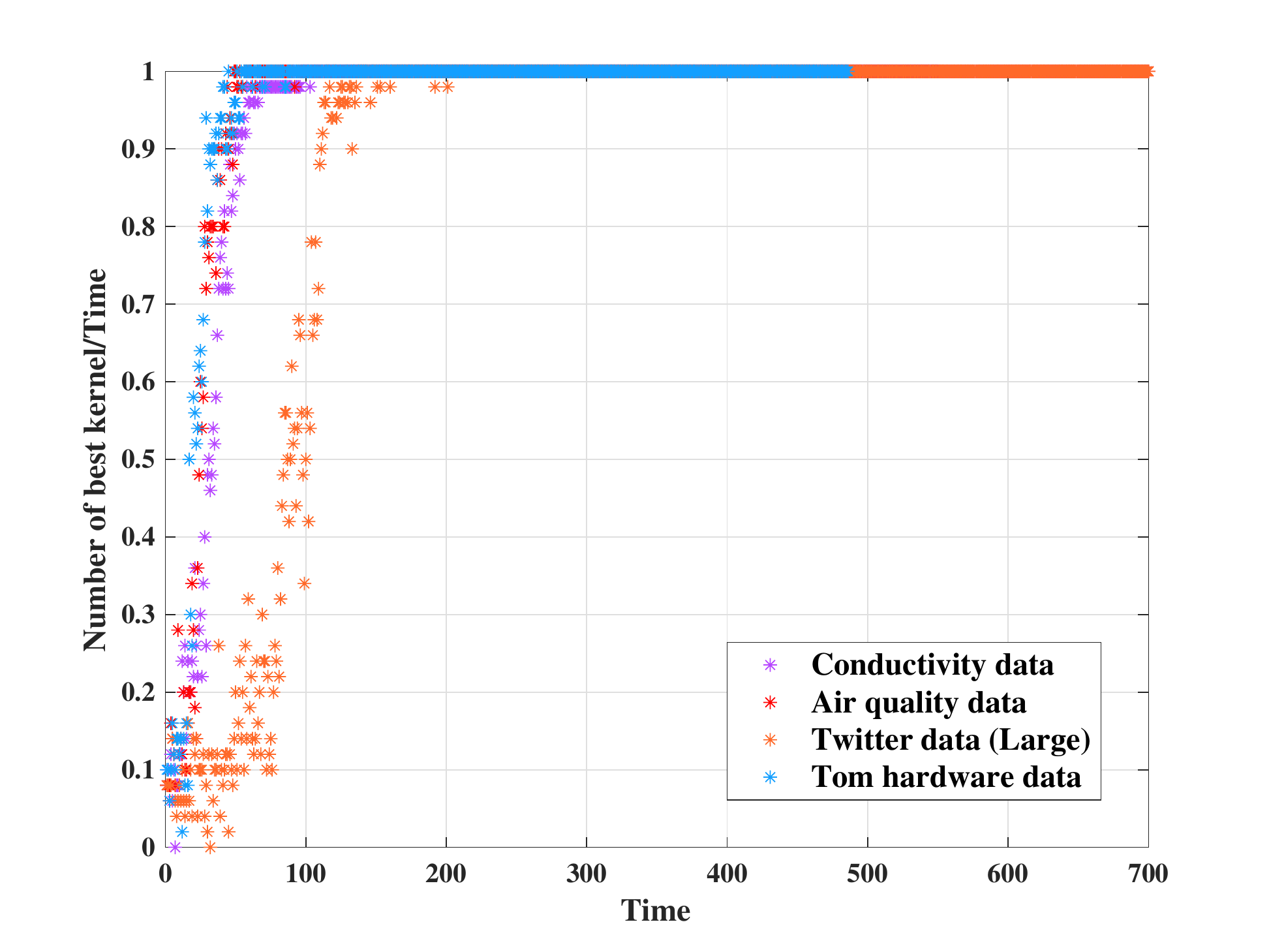}}
\caption{Convergence speed of a randomized MK-OFL by measuring the fraction of the best kernel index out of 50 trials. }
\label{fig:conv}
\end{figure}

\begin{figure*}[!h]
\centering
\subfigure[Twitter data]{
\includegraphics[width=0.34\linewidth]{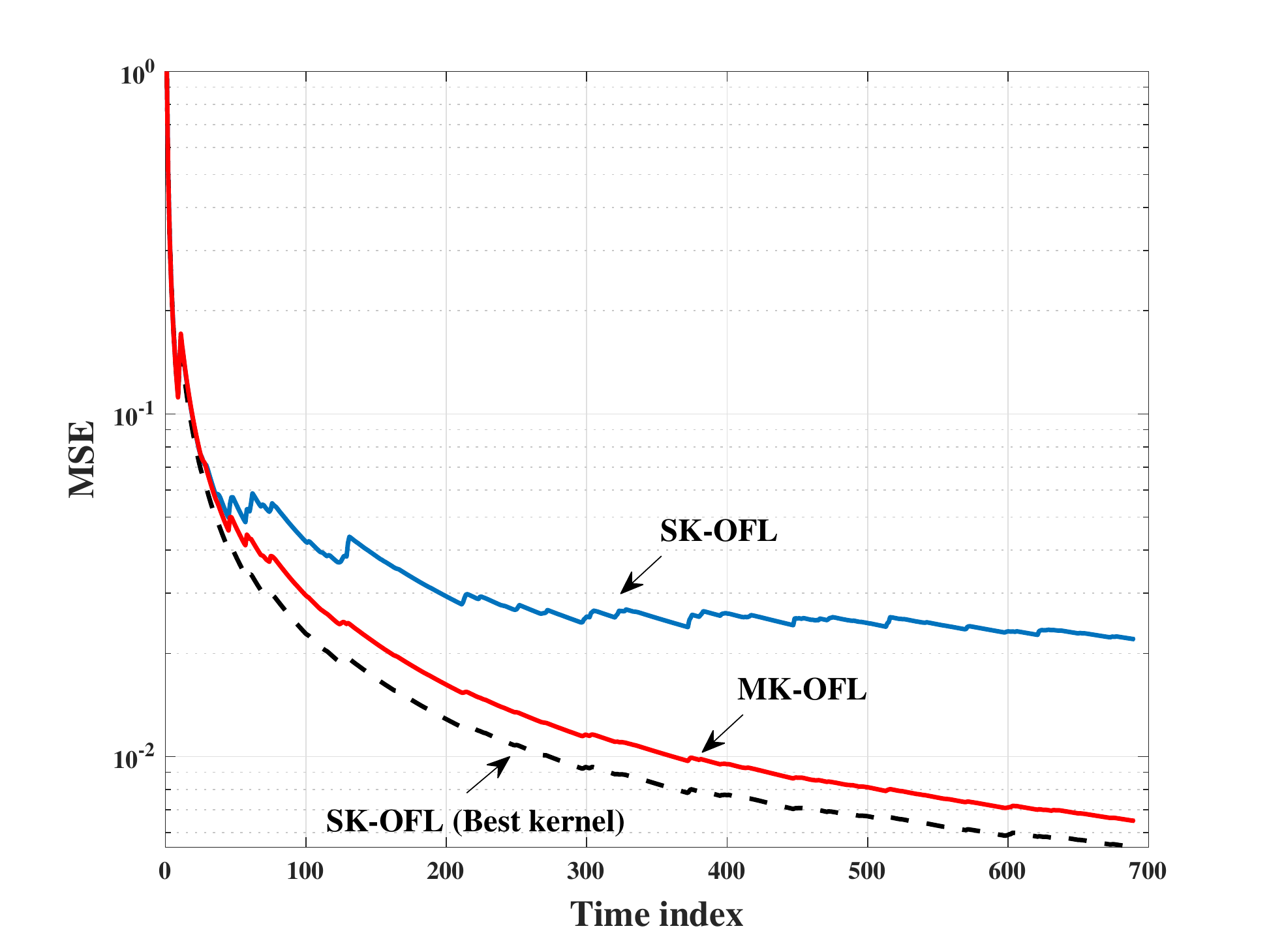}
}
\hspace{-0.7cm}\centering
\subfigure[Twitter data (Large)]{
\includegraphics[width=0.34\linewidth]{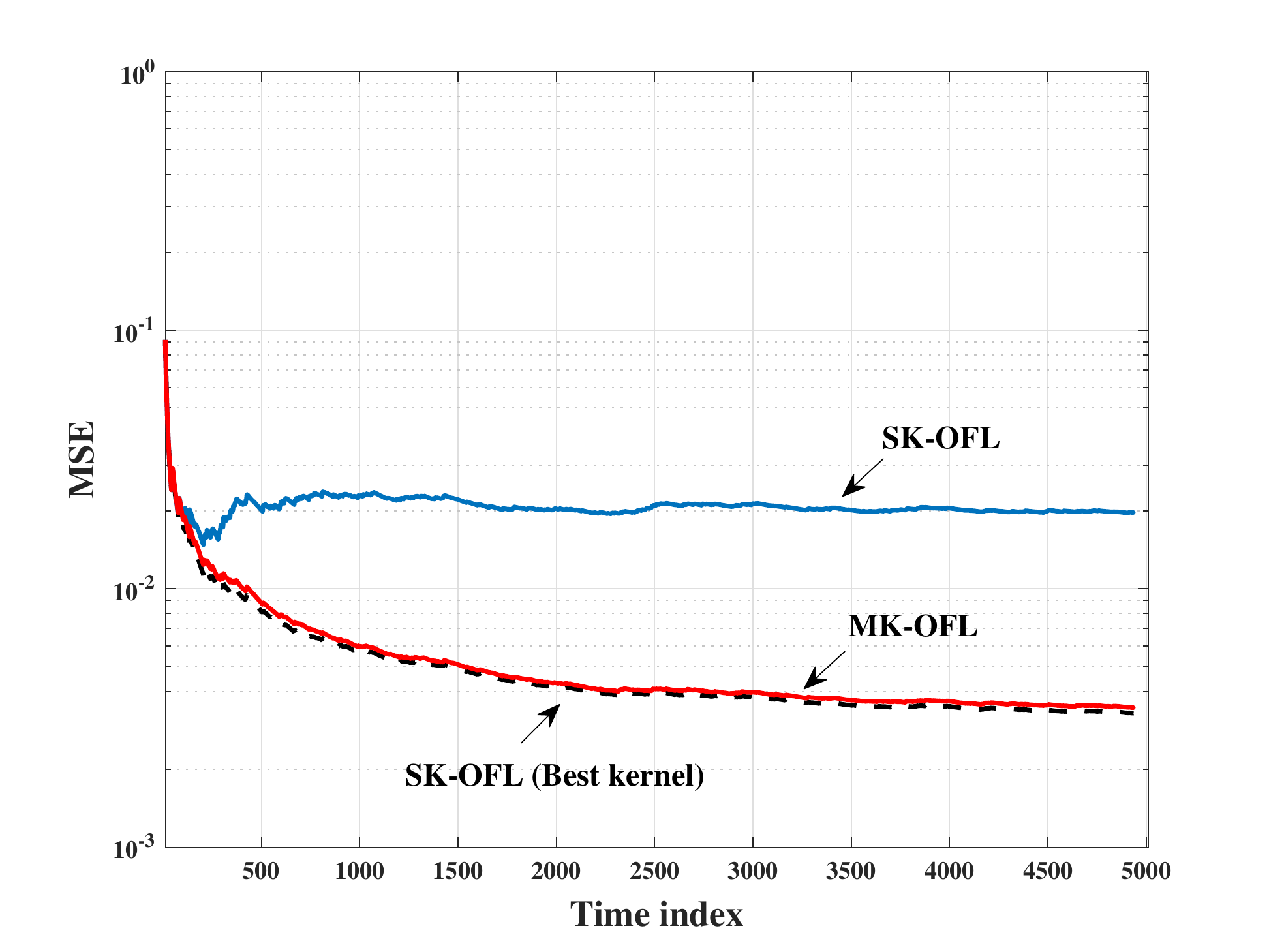}
}
\hspace{-0.7cm}\centering
\subfigure[Tom's hardware data]{
\includegraphics[width=0.34\linewidth]{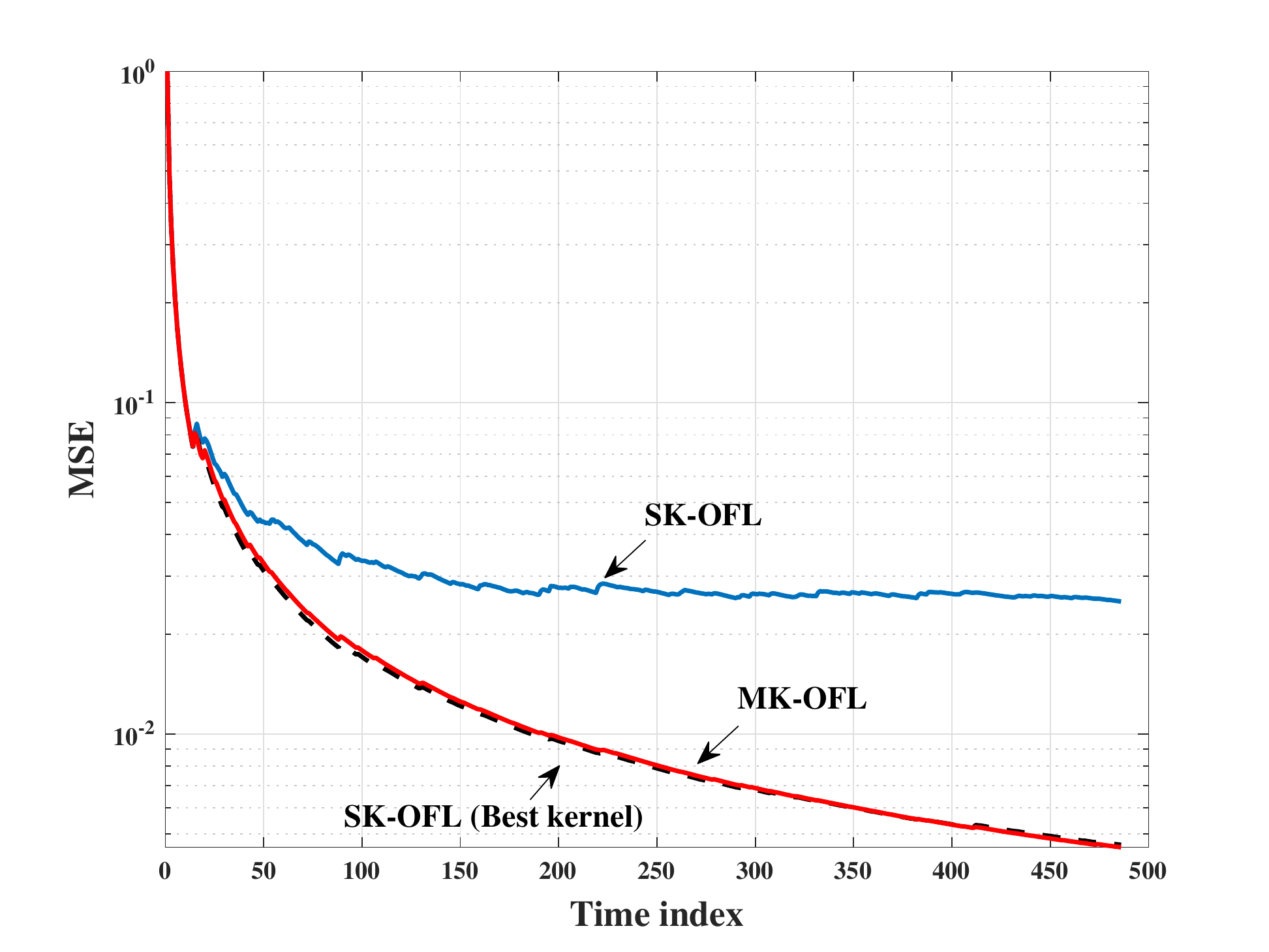}
}
\hspace{-0.7cm}\centering
\subfigure[Conductivity data]{
\includegraphics[width=0.34\linewidth]{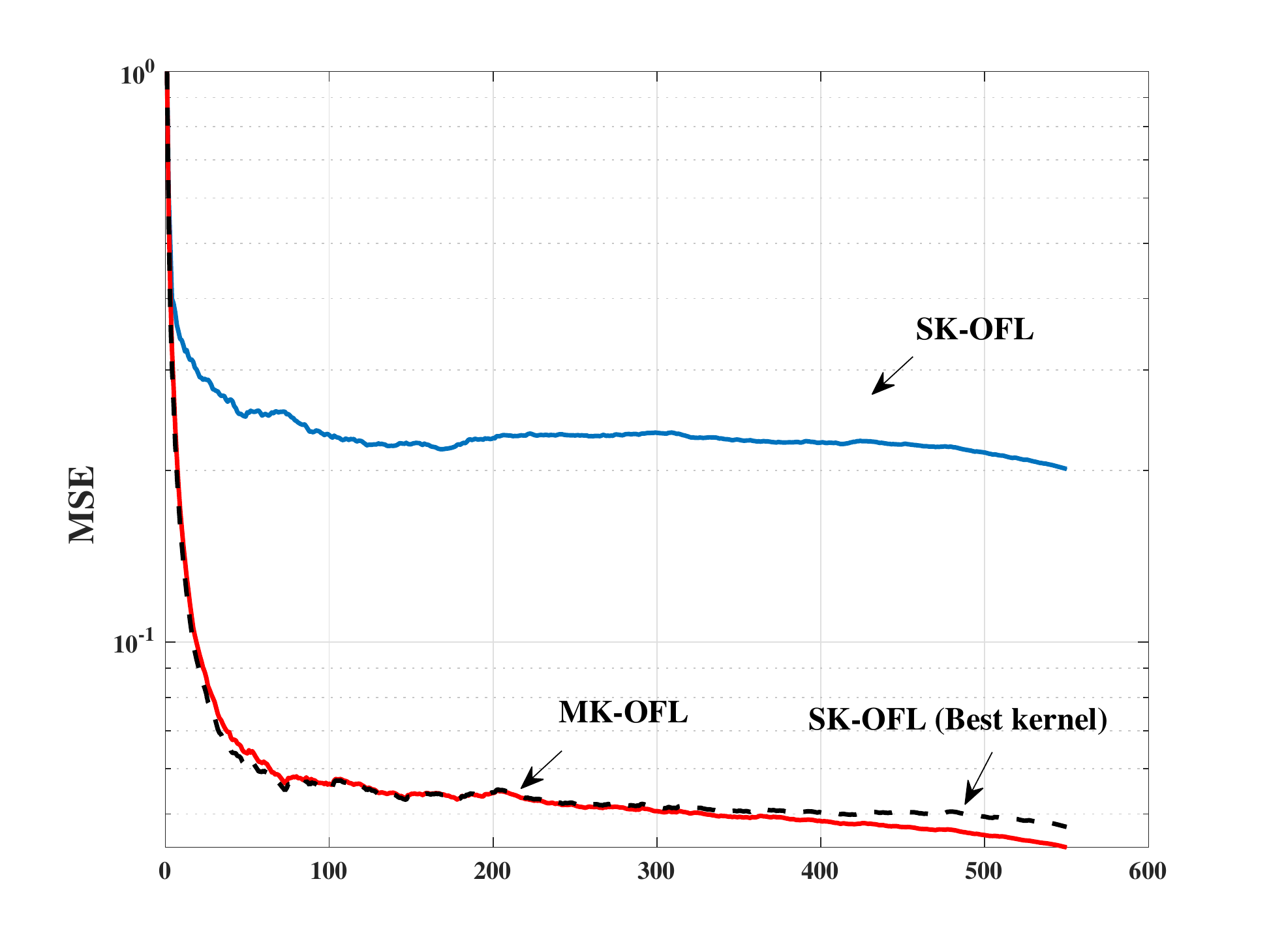}
}
\hspace{-0.7cm}\centering
\subfigure[Wave data]{
\includegraphics[width=0.34\linewidth]{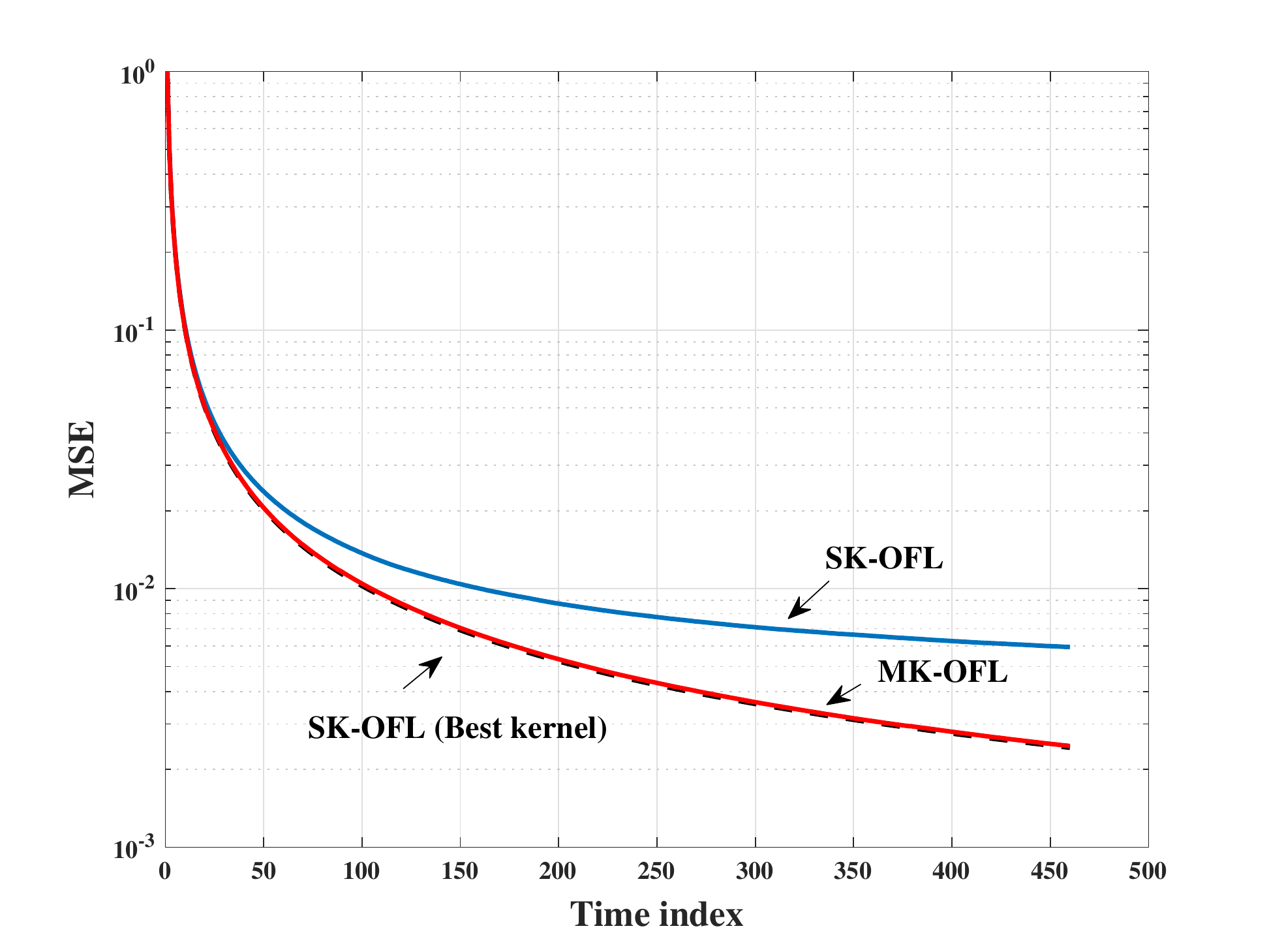}
}
\hspace{-0.7cm}\centering
\subfigure[Naval propulsion plant data]{
\includegraphics[width=0.34\linewidth]{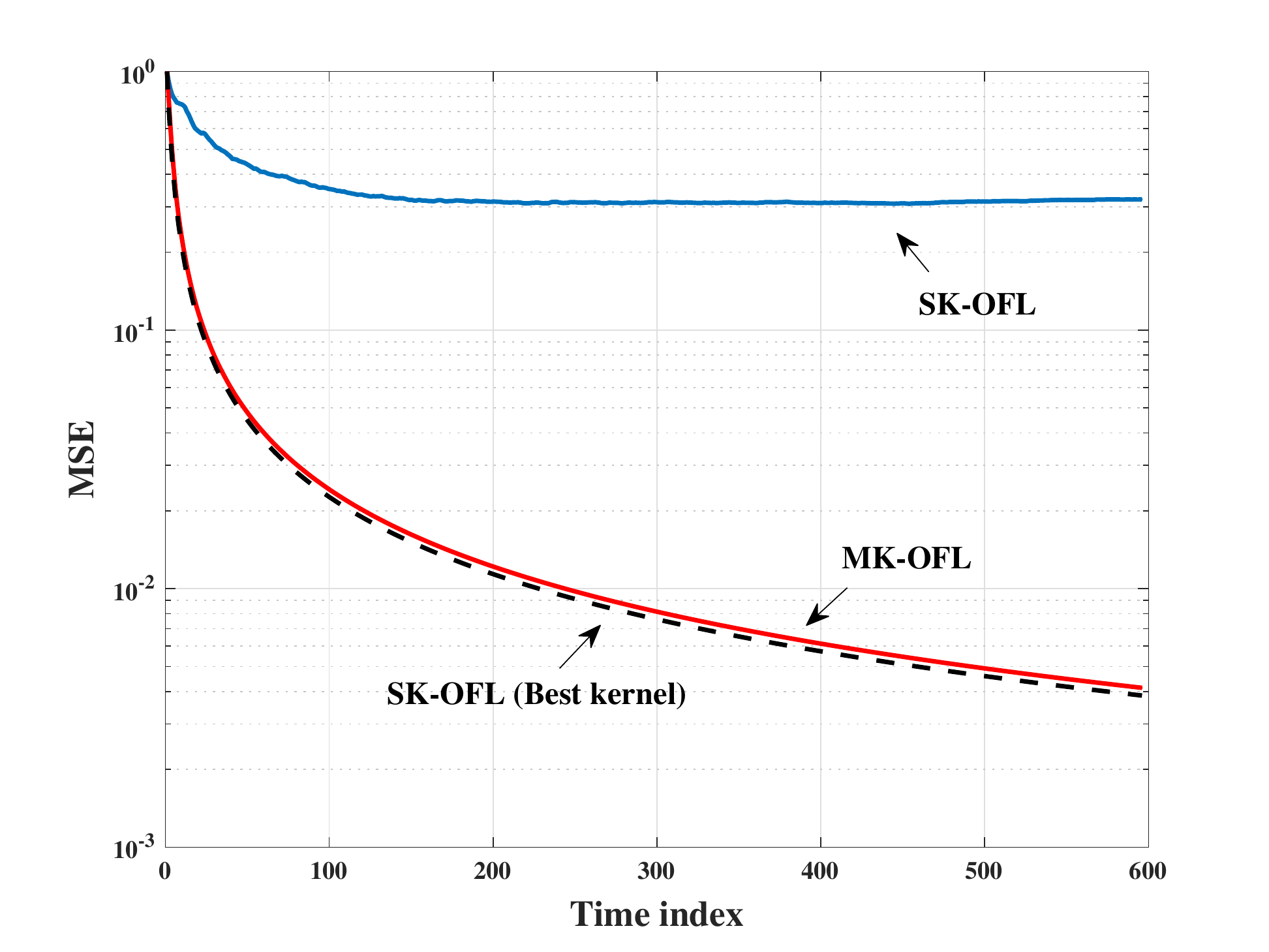}
}
\caption{Comparisons of MSE performances of MK-OFL and SK-OFL in {\bf online regressions tasks}. }
\label{fig:MSEperformance_OR}
\end{figure*}


\begin{figure*}[!h]
\centering
\subfigure[Air quality data]{
\includegraphics[width=0.34\linewidth]{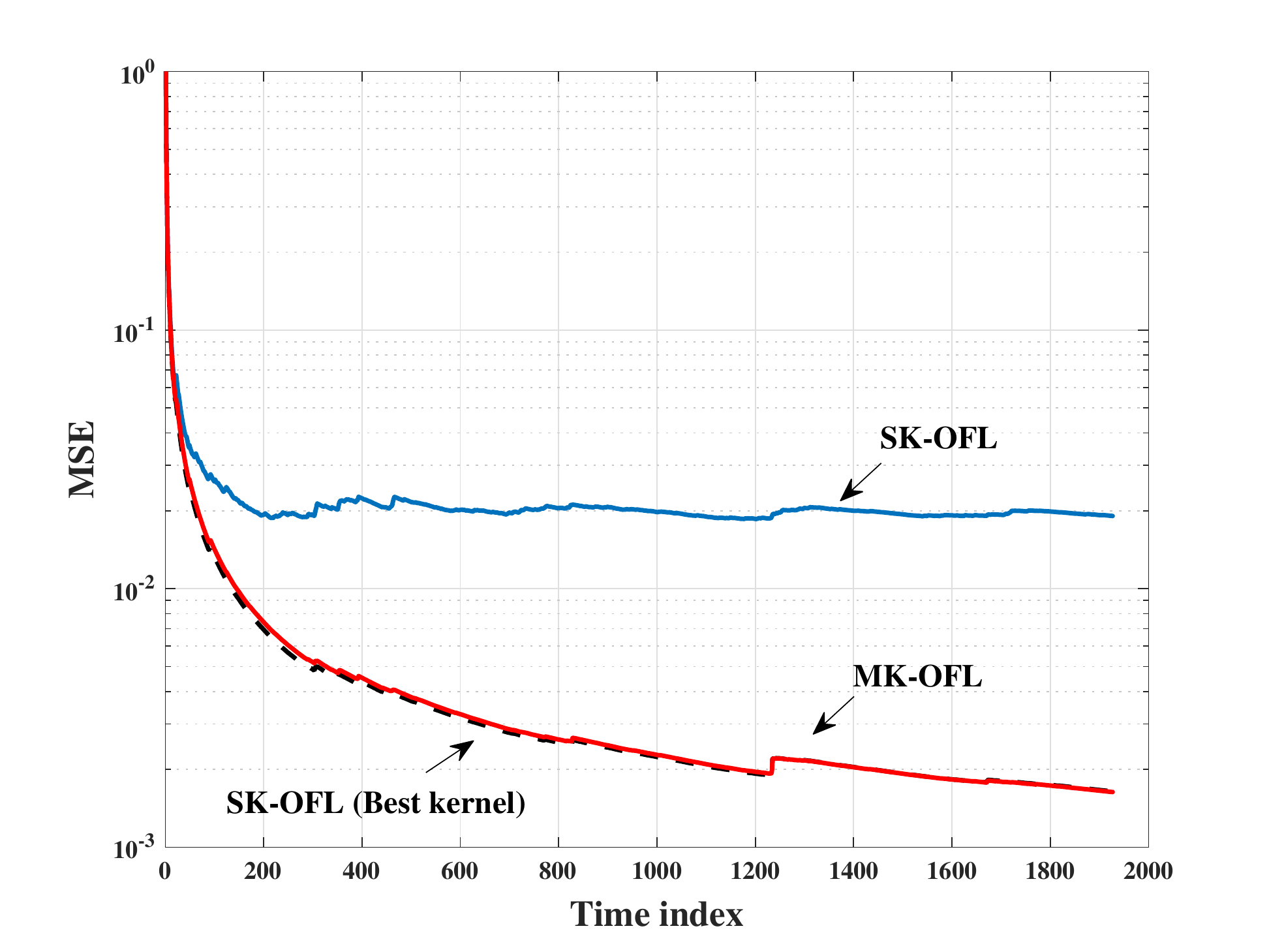}
}
\hspace{-0.7cm}\centering
\subfigure[Traffic data]{
\includegraphics[width=0.34\linewidth]{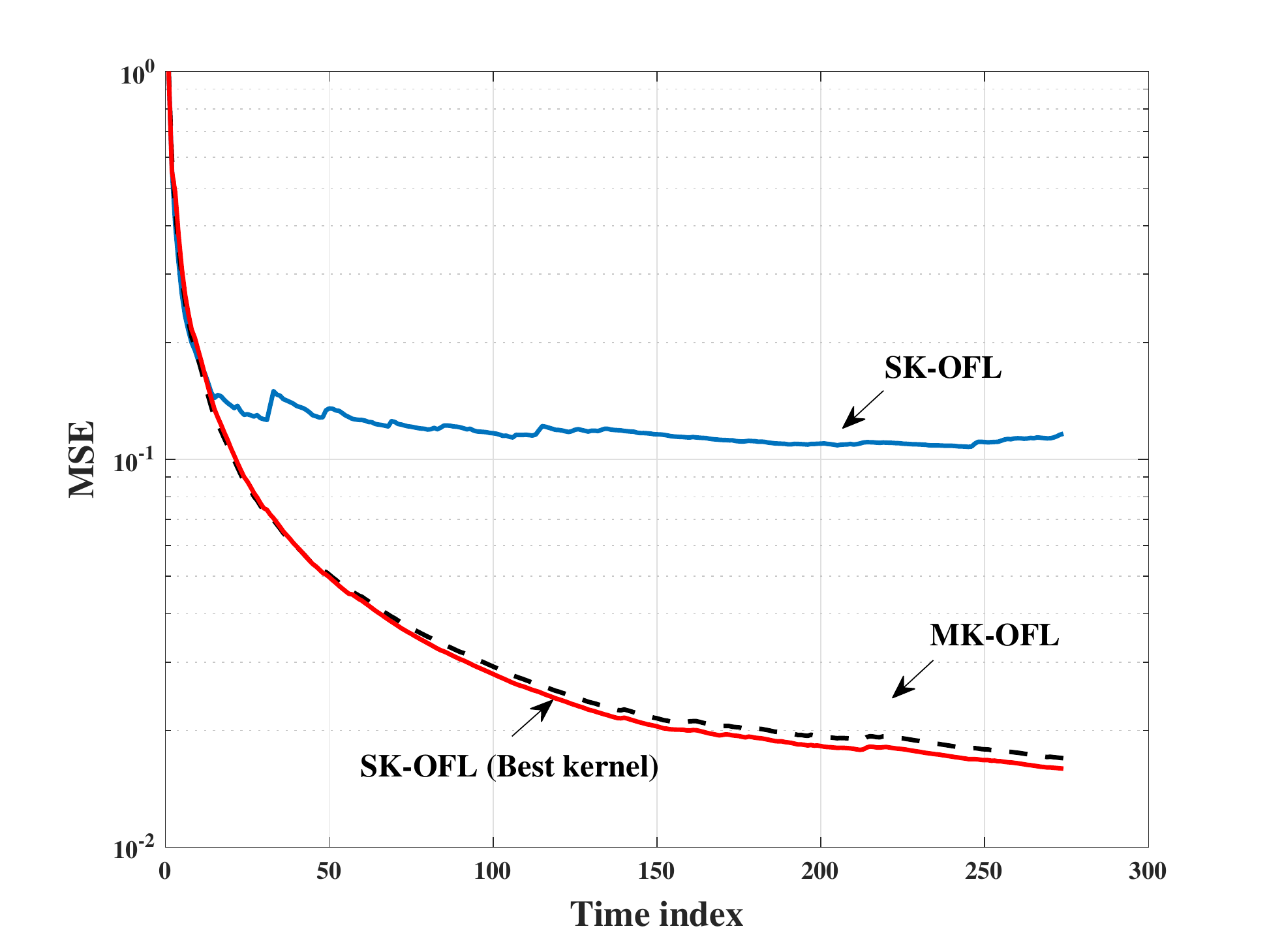}
}
\hspace{-0.7cm}\centering
\subfigure[Temperature data]{
\includegraphics[width=0.34\linewidth]{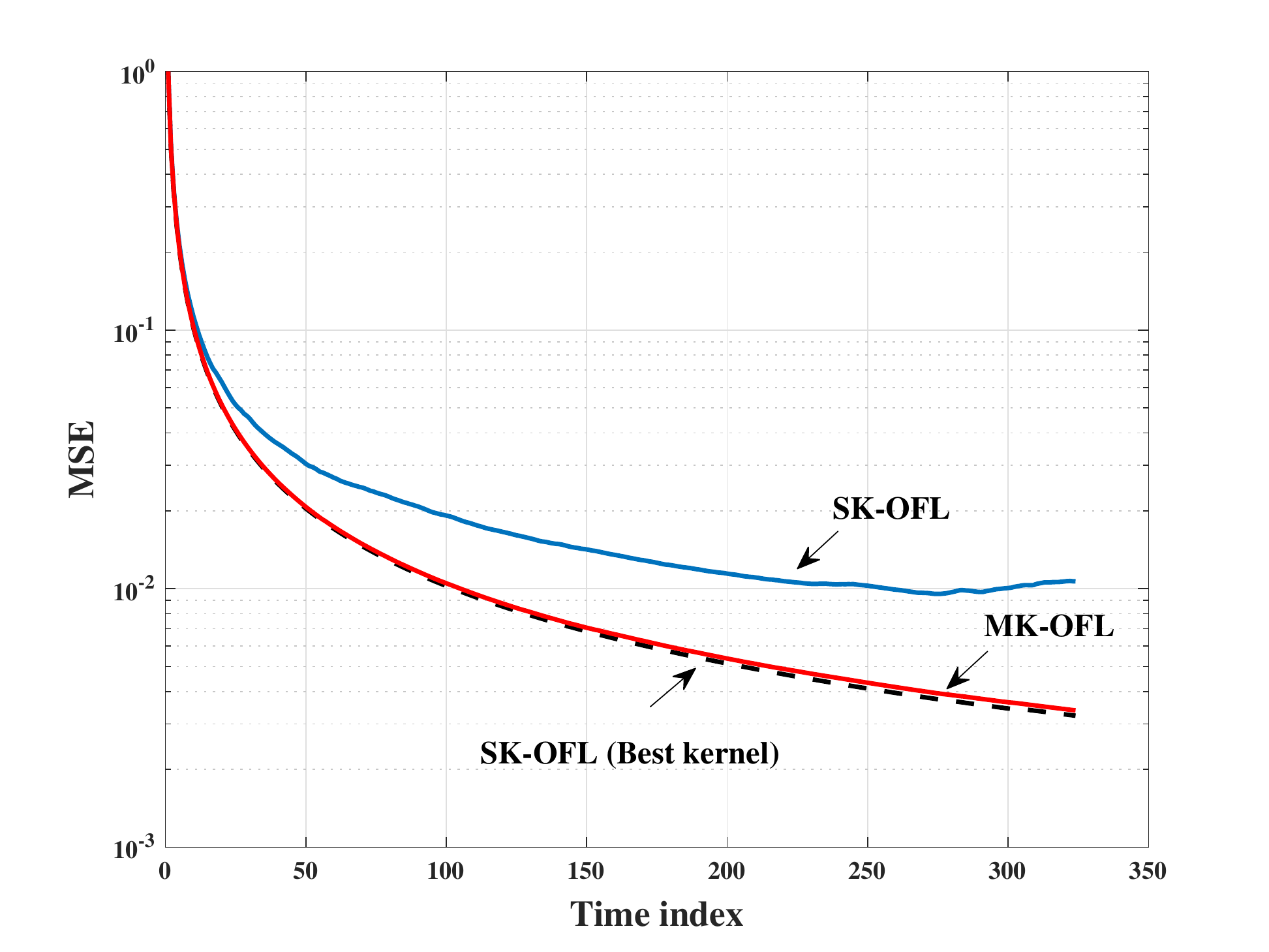}
}
\caption{Comparisons of MSE performances of MK-OFL and SK-OFL in {\bf time-series prediction tasks}.}
\label{fig:MSEperformance_TP}
\end{figure*}

First of all, we demonstrate our analytical result in Theorem~\ref{thm1} with the real-world datasets. The corresponding numerical results are illustrated in Fig.~\ref{fig:conv}, where the fraction of selecting the best kernel $p^{\star} \in [P]$ out of 50 trials is evaluated. It is clearly shown that after a certain time (called a convergence time), our randomized algorithm selects the best kernel out of $P=11$ kernels with probability 1. Furthermore, the convergence time is extremely fast. Because of this, MK-OFL can yield superior performance with a not-so-large number of incoming data (e.g., $T\leq 1000$) as well as an asymptotic case.

We next verify the effectiveness of the proposed MK-OFL in various online learning tasks. Fig.~\ref{fig:MSEperformance_OR} shows the MSE results in online regression tasks with real-world datasets in Section~\ref{subsec:OR}. We observe that MK-OFL almost achieves the accuracy of the best kernel function in all the test datasets. Since the best kernels are generally different according to datasets, it is not likely to determine the best kernel at the beginning of an online learning process (i.e., with no knowledge of dataset). As a consequence, when an suitable kernel is preselected, SK-OFL can deteriorate the accuracy (see Fig.~\ref{fig:MSEperformance_OR}). Especially in Fig.~\ref{fig:MSEperformance_OR} (d) and (f), MKL-OFL and SK-OFL (with an unsuitable kernel) have a non-trivial performance gap. This situation can be happened in numerous real-world applications. Similarly, the MSE performances of MK-OFL in time-series prediction tasks are provided in Fig.~\ref{fig:MSEperformance_TP}. These results also manifest the advantage of using multiple kernels, achieving the performance of the best kernel function and outperforming SK-OFL considerably. From the experimental results in Fig.~\ref{fig:MSEperformance_OR} and Fig.~\ref{fig:MSEperformance_TP}, one can expect that MK-OFL yields an outstanding performance for any real-world application with a sufficiently large number of kernels. This is a natural consequence as the probability of the adequate kernel belongs to a set of $P$ kernels becomes higher as $P$ grows. In contrast, using multiple similar kernels might degrade the performance since it can decrease the convergence time especially when there exist several kernels similar to the best one. In this regard, one interesting future work is to construct $P$ kernels so that they are distinct enough. With a sufficiently large $P$, this construct can guarantee a fast convergence time as well as increase the probability containing the adequate kernel in the kernel dictionary. This is left of an interesting future work.

\subsection{Online Regression Tasks}\label{subsec:OR}
For the experiments of online regression tasks, the following real datasets from UCI Machine Learning Repository are considered:
\vspace{0.2cm}
\begin{itemize}
    \item {\bf Twitter \cite{Kawala2013}:} Data contains buzz events from Twitter, where each attributes are used to predict the popularity of a topic. Higher value indicates more popularity. The larger dataset (termed {\bf Twitter(L)}) is also considered.
    \item {\bf Tom's hardware \cite{Kawala2013}:} Data  consists of samples acquired from a forum, where each features represents such as the number of times a content is displayed to visitors. The task is to predict the average number of display about on a certain topic.
    \item {\bf Conductivity \cite{Hamidieh}:} This dataset contains samples of extracted from superconductors, where each feature represents critical information to construct superconductor such as density and mass of atoms. The goal is to predict the critical temperature to create superconductor.
    \item {\bf Wave \cite{Neshat}:} This data contains the samples consisting of positions and absorbed power obtained from wave energy converters (WECs) in four real wave scenarios from the southern coast of Australia. The goal is to predict total power energy of the farm.
    \item {\bf Naval propulsion plants \cite{Coraddu2014}:} This dataset has been obtained from Gas Turbine plant. Dataset contains samples with 16 features such as ship speed and fuel flow. The goal is to determine turbine decay state coefficient.
\end{itemize}

\subsection{Time-series Prediction Tasks}\label{subsec:TP}

We consider time-series prediction tasks which estimate the future values in online fashion. As considered in the centralized counterpart \cite{hong2020active}, the famous time-series prediction method called Autogressive (AR) model is considered. An ${\rm AR}(s)$ model predicts the future value $\yv_t$ assuming the linear dependency on its $s$ values, i.e.,
\begin{equation}
    y_t = \sum_{i=1}^s \gamma_i y_{t-i} + n_t,
\end{equation} where $\gamma_i$ denotes the weight for $y_{t-i}$ and $n_t$ denotes a Gaussian noise at time $t$. Based on this, the RF-based kernelized ${\rm AR}(s)$ model, which can explore a nonlinear dependency, is introduced in \cite{hong2020active}, given as
\begin{align}\label{eq:model}
    y_t = f_t(\xv_t) + n_t,
\end{align} where $\xv_{t} = [y_{t-1},...,y_{t-s}]^{\trasp}$. The proposed MKL-OFL aims at learning $f_t(\cdot)$ with a parameterized model $\hat{f}_{t}(\xv)=\hat{\wv}_{t}\zv_{\hat{p}_{t}}(\xv)$. The proposed algorithm is tested with the following univariate time-series datasets from UCI Machine Learning Repository:
\vspace{0.2cm}
\begin{itemize}
    \item {\bf Air quality \cite{DeVito2008}:} Data includes samples, of which features include hourly response from an array of chemical sensors embedded in a city of Italy. The goal is to predict the concentration of polluting chemicals in the air.
    \item {\bf Traffic \cite{Dua:2019}:} This dataset contains the time-series traffic data obtained from Minneapolis Department of Transportation in US. Data is collected from hourly interstate 94 Westbound traffic volume for MN DoT ATR station 301, roughly midway between Minneapolis and St Paul, MN.
    \item {\bf Temperature \cite{Dua:2019}:} This dataset consists of the time-series temperature data obtained from Minneapolis Department of Transportation in US. Data is collected from hourly interstate 94 Westbound temperature for MN DoT ATR station 301, roughly midway between Minneapolis and St Paul, MN.
\end{itemize}

\section{Conclusion}\label{sec:conclusion}

We have proposed a novel online federated learning (OFL) framework with multiple kernels. The proposed method is dubbed multiple kernel-based OFL (MK-OFL). In the proposed MK-OFL, one kernel index is random selected from a predetermined set of $P$ kernels at every time according to a carefully designed probability distribution. Also, the chosen index quickly converges to the best kernel in hindsight. Thus, our algorithm indeed exploits the advantage of using multiple kernels while having the same communication overhead with a single kernel-based OFL (SK-OFL). The asymptotic optimality of MK-OFL is proved using a martingale argument. Beyond the asymptotic analysis, we demonstrated the effectiveness of our algorithm in various online learning tasks with real-world datasets, which suggests practicality. One interesting future work is to extend MK-OFL into a wireless OFL framework, in which a distributed optimization at edge nodes and an optimal estimation at a server will be incorporated. Another interesting future work is to construct the so-called {\em collaborative OFL} by integrating collaborating learning with OFL so as to enable edge nodes to engage in OFL without directly connecting a server.



\appendix

\section{Proof of Lemma 1}\label{app:proof1}

Given the parameters $\hat{p}_{t} \in [P]$ and $\hat{\wv}_{t}$, the OGD update for any fixed $p \in [P]$ is given as
\begin{equation}
    \tilde{\wv}_{k,t+1}^{p} = \hat{\wv}_{k,t}^{p} - \eta_{\ell}\nabla\Lc((\hat{\wv}_{k,t}^p)^{\trasp}\zv_{p}(\xv_{k,t}),y_{k,t}),\label{proof:update}
\end{equation}where
\begin{equation}\label{eq:lem1-0}
    \hat{\wv}_{k,t}^p = 
    \begin{cases}
    \tilde{\wv}_{k,t}^p & \mbox{ if } \hat{p}_{t} \neq p\\
    \hat{\wv}_{t}=\frac{1}{K}\sum_{k=1}^{K}\tilde{\wv}_{k,t}^p & \mbox{ if } \hat{p}_{t} = p.
    \end{cases}
\end{equation} Our analysis is based on the regret analysis of OGD but carefully considering the above cases. From \eqref{proof:update} and following the analysis of OGD, we have that for any $k \in \Vc$:
\begin{align}
    &\|\tilde{\wv}_{k,t+1}^p  - \wv_{p}^{\star}\|^2 \nonumber\\
    &= \|\hat{\wv}_{k,t}^p - \eta_{\ell} \nabla\Lc((\hat{\wv}_{k,t}^p)^{\trasp}\zv_{p}(\xv_{k,t}),y_{k,t})- \wv_{p}^{\star}\|^2 \nonumber\\
    &=\|\hat{\wv}_{k,t}^p  - \wv_{p}^{\star}\|^2 + \eta_{\ell}^2\|\nabla\Lc((\hat{\wv}_{k,t}^p)^{\trasp}\zv_{p}(\xv_{k,t}),y_{k,t})\|^2\nonumber\\ &-2\eta_{\ell}\nabla\Lc((\hat{\wv}_{k,t}^p)^{\trasp}\zv_{p}(\xv_{k,t}),y_{k,t})^{\trasp}(\hat{\wv}_{k,t}^p-\wv_{p}^{\star}).
\end{align} Foe ease of exposition, throughout the proof, we let
\begin{equation}
    \nabla_{k,t}^p \eqdef \nabla\Lc((\hat{\wv}_{k,t}^p)^{\trasp}\zv_{p}(\xv_{k,t}),y_{k,t}).
\end{equation} According to the two cases in \eqref{eq:lem1-0}, we can get:

i) For $\hat{\wv}_{k,t}^{p} = \tilde{\wv}_{k,t}^p$ (i.e., $p\neq \hat{p}_t$), we have:
\begin{align}
    &\|\tilde{\wv}_{k,t+1}^p  - \wv_{p}^{\star}\|^2= \|\tilde{\wv}_{k,t}^p  - \wv_{p}^{\star}\|^2 \nonumber\\
    &\;\;\;\;\;\;\;\;\;\;\; + \eta_{\ell}^2\|\nabla_{k,t}^p\|^2-2\eta_{\ell}(\nabla_{k,t}^p)^{\trasp}(\hat{\wv}_{k,t}^p-\wv_{p}^{\star}).\label{eq:lem1-11}
\end{align} 

ii) For $\hat{\wv}_{k,t}^{p}=\frac{1}{K}\sum_{k=1}^{K}\tilde{\wv}_{k,t}^p$ (i.e., $p = \hat{p}_t$), we have:
\begin{align}
   &\sum_{k \in [K]} \|\tilde{\wv}_{k,t+1}^p  - \wv_{p}^{\star}\|^2
    \stackrel{(a)}{\leq} \sum_{k \in [K]}\|\tilde{\wv}_{k,t}^p  - \wv_{p}^{\star}\|^2\nonumber\\ 
    &+\eta_{\ell}^2\sum_{k \in [K]}\|\nabla_{k,t}^p\|^2 -2\eta_{\ell}\sum_{k \in [K]}(\nabla_{k,t}^p)^{\trasp}(\hat{\wv}_{k,t}^p-\wv_{p}^{\star}),\label{eq:lem1-12}
\end{align} where (a) is due to the following fact:
\begin{align*}
    &\sum_{k \in [K]}\|\hat{\wv}_{k,t}^p  - \wv_{p}^{\star}\|^2=\frac{1}{K^2}\sum_{k \in [K]}\left\|\sum_{\ell \in [K]}\tilde{\wv}_{\ell,t}^p  - \wv_{p}^{\star}\right\|^2\\
    &\leq\frac{1}{K}\sum_{k \in [K]}\|\tilde{\wv}_{k,t}^p - \wv_{p}^{\star}\|^2\leq \sum_{k \in [K]}\|\tilde{\wv}_{k,t}^p - \wv_{p}^{\star}\|^2.
\end{align*} 
Leveraging the convexity of a loss function, we obtain that for any $k \in \Vc$: 
\begin{align}
    &\Lc((\hat{\wv}_{k,t}^p)^{\trasp}\zv_{p}(\xv_{k,t}),y_t) - \Lc((\wv_{p}^{\star})^{\trasp}\zv_{p}(\xv_{k,t}),y_t)\nonumber\\
    &\leq (\nabla_{k,t}^p)^{\trasp}(\hat{\wv}_{k,t}^p-\wv_{p}^{\star}). \label{eq:lem1-2}
\end{align} Plugging \eqref{eq:lem1-2} into \eqref{eq:lem1-11} and \eqref{eq:lem1-12} separately, and combining these two cases, we can get
\begin{align}
     &\sum_{k=1}^{K}\Lc((\hat{\wv}_{k,t}^p)^{\trasp}\zv_{p}(\xv_{k,t}),y_t) - \Lc((\wv_{p}^{\star})^{\trasp}\zv_{p}(\xv_{k,t}),y_t) \nonumber\\
     &\leq \sum_{k=1}^{K}\frac{\|\tilde{\wv}_{k,t}^p  - \wv_{p}^{\star}\|^2-\|\tilde{\wv}_{k,t+1}^p  - \wv_{p}^{\star}\|^2}{2\eta_{\ell}}\nonumber\\
     &+\frac{\eta_{\ell}}{2}\sum_{k=1}^{K}\|\nabla_{k,t}^p\|^2.\label{eq:lem1-f}
\end{align} Summing \eqref{eq:lem1-f} over $t=1,...,T$, we obtain that for any fixed $p \in [P]$,
\begin{align}
     &\sum_{t=1}^{T}\sum_{k=1}^{K} \Lc((\tilde{\wv}_{k,t}^p)^{\trasp}\zv_{p}(\xv_{k,t}),y_t) - \Lc((\wv_{p}^{\star})^{\trasp}\zv_{p}(\xv_{k,t}),y_t)\nonumber\\
      &\stackrel{(a)}{\leq} \sum_{t=1}^{T}\sum_{k=1}^{K} \frac{\|\tilde{\wv}_{k,1}^p  - \wv_{p}^{\star} \|^2}{2\eta_{\ell}}+ \frac{\eta_{\ell}}{2}\sum_{t=1}^{T}\sum_{k=1}^{K}\|\nabla_{k,t}^p\|^2\nonumber\\
    &\stackrel{(b)}{\leq} \frac{KC^2}{2\eta_{\ell}} + \frac{\eta_{\ell} K L^2 T}{2},
\end{align} where (a) is due to the telescoping sum and (b) is from the Assumption 2 and Assumption 3. This completes the proof of Lemma 1.

\section{Proof of Lemma 2}\label{app:proof2}
Recall that
\begin{equation}
    L_{t}^p = \exp\left(-\eta_g\sum_{k=1}^K \Lc((\hat{\wv}_{k,t}^{p})^{\trasp}\zv_{p}(\xv_{k,t}), y_{k,t}) \right).
\end{equation} Also, to simplify the expressions, we let
\begin{equation}
    \hat{f}_{k,t}^p(\xv) \eqdef (\hat{\wv}_{k,t}^p)^{\trasp}\zv_{p}(\xv).
\end{equation} Then, the proof will be completed by comparing the upper and lower bounds of $\zeta$, which is defined below:
\begin{align*}
\zeta = \sum_{t=1}^T \log\left(\sum_{p=1}^{P}\hat{\qv}_{t}(p)\exp\left(-\eta_g\sum_{k=1}^K \Lc(\hat{f}_{k,t}^{p}(\xv_{k,t}),y_{k,t})\right)\right).
\end{align*} We first derive the upper bound on $\zeta$ such as
\begin{align*}
    \zeta &= \sum_{t=1}^{T}\log\left(\EE_{I_t}\left[\exp\left(-\eta_g\sum_{k=1}^K \Lc(\hat{f}_{k,t}^{I_t}(\xv_{k,t}),y_{k,t}) \right)
    \right]\right)\\
    &\stackrel{(a)}{\leq} -\eta_g\sum_{t=1}^{T}\sum_{k=1}^{K}\EE_{I_t}\left[\Lc(\hat{f}_{k,t}^{I_t}(\xv_{k,t}),y_{k,t})\right] + \frac{\eta_g^2 KT}{8}\\
    &\stackrel{(b)}{\leq} -\eta_g\sum_{t=1}^{T}\sum_{k=1}^{K}\Lc\left(\EE_{I_t}\left[\hat{f}_{k,t}^{I_t}(\xv_{k,t})\right],y_{k,t}\right) + \frac{\eta_g^2 KT}{8},
\end{align*} where (a) follows the Hoeffding's inequality with the bounded random variable $\Lc(\hat{f}_{k,t}^{I_t}(\xv_{k,t}),y_{k,t})\in [0,1]$ and (b) is from Assumption 1 (i.e., the convexity of a loss function). We next derive the lower bound on $\zeta$ as follows:
\begin{align*}
   &\zeta=\sum_{t=1}^{T} \log\left(\sum_{p=1}^{P}\hat{\qv}_{t}(p)\exp\left(-\eta_g\sum_{k=1}^{K}\Lc(\hat{f}_{k,t}^{p}(\xv_{k,t}),y_{k,t})\right)\right)\\
    &\stackrel{(a)}{=}\sum_{t=1}^{T} \log\left( \sum_{p=1}^{P}\bar{\qv}_{t}(p)\exp\left(-\eta_g\sum_{k=1}^{K}\Lc(\hat{f}_{k,t}^{p}(\xv_{k,t}),y_{k,t})\right)\right) \nonumber\\
    &\;\;\;\;\; +\sum_{t=1}^{T} \log\left(\frac{\sum_{p=1}^{P}\hat{\qv}_{t}(p) L_{t}^{p}}{\sum_{p=1}^{P}\bar{\qv}_{t}(p)L_{t}^p}\right)\\
    &\stackrel{(b)}{=}\sum_{t=1}^{T}\log\left(\frac{\sum_{p=1}^{P}\bar{\mv}_{t+1}(p)}{\sum_{p=1}^{P}\bar{
    \mv}_{t}(p)}\right)  +\sum_{t=1}^{T} \log\left(\frac{\sum_{p=1}^{P}\hat{\qv}_{t}(p) L_{t}^{p}}{\sum_{p=1}^{P}\bar{\qv}_{t}(p)L_{t}^p}\right)\nonumber\\
    &\stackrel{(c)}{=}\log\left(\sum_{p=1}^{P}\bar{\mv}_{T+1}(p)\right) - \log\left(\sum_{p=1}^{P}\bar{\mv}_{1}(p)\right)\nonumber\\
    &\;\;\;\;\; +\sum_{t=1}^{T} \log\left(\frac{\sum_{p=1}^{P}\hat{\qv}_{t}(p) L_{t}^{p}}{\sum_{p=1}^{P}\bar{\qv}_{t}(p)L_{t}^p}\right)
    \end{align*}
    \begin{align*}
    &\geq \log\left(\max_{1\leq p \leq P}\bar{\mv}_{T+1} \right) - \log{P} \nonumber\\
    &\;\;\;\;\; +\sum_{t=1}^{T} \log\left(\frac{\sum_{p=1}^{P}\hat{\qv}_{t}(p) L_{t}^{p}}{\sum_{p=1}^{P}\bar{\qv}_{t}(p)L_{t}^p}\right)\\
    &=-\eta_g \min_{1\leq p \leq P} \sum_{t=1}^T \sum_{k=1}^{K} \Lc(\hat{f}_{k,t}^{p}(\xv_{k,t}),y_{k,t}) - \log{P} \nonumber\\
    &\;\;\;\;\; +\sum_{t=1}^{T} \log\left(\frac{\sum_{p=1}^{P}\hat{\qv}_{t}(p) L_{t}^{p}}{\sum_{p=1}^{P}\bar{\qv}_{t}(p)L_{t}^p}\right),
\end{align*}where (a) follows the definition of $L_t^p$ in the main document,  (b) is from the definitions of $\bar{\qv}_t(p)$ and $\bar{w}_{t}(p)$,  and (c) is due to the telescoping sum. From the upper and lower bounds, we can get
\begin{align*}
    &-\eta_g \min_{1\leq p \leq P} \sum_{t=1}^T \sum_{k=1}^{K} \Lc(\hat{f}_{k,t}^{p}(\xv_{k,t}),y_{k,t}) - \log{P} \nonumber\\
    &+\sum_{t=1}^{T} \log\left(\frac{\sum_{p=1}^{P}\hat{\qv}_{t}(p) L_{t}^{p}}{\sum_{p=1}^{P}\bar{\qv}_{t}(p)L_{t}^p}\right)\\
    &\leq -\eta_g\sum_{t=1}^{T}\sum_{k=1}^{K}\Lc\left(\EE_{I_t}[\hat{f}_{k,t}^{I_t}(\xv_{k,t})],y_{k,t}\right) + \frac{\eta_g^2 K T}{8}.
\end{align*}  By rearranging them, we have:
\begin{align*}
    &\sum_{t=1}^{T}\sum_{k=1}^{K}\Lc\left(\sum_{p=1}^{P}\hat{q}_{t}(p) \left(\hat{\wv}_{k,t}^p\right)^{\trasp}\zv_{p}(\xv_{k,t}),y_{k,t}\right)\nonumber\\
    &- \min_{1\leq p \leq P} \sum_{t=1}^T \sum_{k=1}^{K} \Lc\left( (\hat{\wv}_{k,t}^p)^{\trasp}\zv_{p}(\xv_{k,t}),y_{j,t}\right)\nonumber\\
    &\leq \frac{1}{\eta_g} \log{P} + \frac{ \eta_g  K T}{8}+\sum_{t=1}^{T} \log\left(\frac{\sum_{p=1}^{P}\hat{\qv}_{t}(p)L_{t}^p }{\sum_{p=1}^{P}\bar{\qv}_{t}(p)L_{t}^{p}}\right).
\end{align*} Also, we have the following inequality:
\begin{align*}
    &\sum_{t=1}^{T} \log\left(\frac{\sum_{p=1}^{P}\hat{\qv}_{t}(p) L_{t}^{p}}{\sum_{p=1}^{P}\bar{\qv}_{t}(p)L_{t}^p}\right)\nonumber\\
    &\;\;\;\;\;\;\;\;\;\;\; \stackrel{(a)}{\leq} \sum_{t=1}^{T} \log\left(\frac{\sum_{p=1}^{P}\hat{\qv}_{t}(p)}{\sum_{p=1}^{P}\bar{\qv}_{t}(p)}\right) + \eta_g K T\\
    &\;\;\;\;\;\;\;\;\;\;\; \stackrel{(b)}{=} \eta_g K T,
\end{align*} where (a) is due to the fact that $\exp(-K\eta_g) \leq L_{t}^{p} \leq  1$ from Assumption 1 and (b) is from that $\sum_{p=1}^{P}\hat{\qv}_{t}(p) = \sum_{p=1}^{P}\bar{\qv}_{t}(p) = 1$. From (3) and (4), the proof is completed.

\section{Proof Lemma 3}\label{app:proof3}

For the proof, we define a random variable $X_{k,t}$ as
\begin{align*}
    X_{k,t} &= \Lc(W_{t}^{\trasp}\zv_{I_t}(\xv_{k,t}),y_{k,t}) \nonumber\\
    &- \sum_{p=1}^{P} \hat{\qv}_{t}(p)\Lc((\hat{\wv}_{k,t}^{p})^{\trasp}\zv_p(\xv_{k,t}),y_{k,t}).
\end{align*} Here, one can think that the fixed values $\hat{\wv}_{k,t}^{p}$ and $\hat{\qv}_{t}(p)$ are obtained as a consequence of random variables $I_1,...,I_{t-1}$. Specifically, let $\Fc_{t}=\sigma(I_1,I_2,...,I_t)$ be the smallest sigma algebra such that $I_1,I_2,...,I_t$ is measurable. Then, $\{\Fc_t: t=1,...,T\}$ is filtration and $X_{k,t}$ is $\Fc_t$-measurable. Note that condition on $\Fc_{t-1}$, $\hat{\qv}_{t}(p)$ and $\hat{\wv}_{k,t}^{p}$ are fixed and $I_t$ is only random variable. Leveraging this fact, we have:
\begin{equation*}
    \EE[X_{j,t}|\Fc_{t-1}]=0,
\end{equation*} because
\begin{align*}
    &\EE\left[\Lc\left(W_{t}^{\trasp}\zv_{I_t}(\xv_{k,t}),y_{k,t}\right)\Big|\Fc_{t-1}\right] \\
    &\;\;\;\;\;\;\;\;\;\;\;\;\;\;\;\; = \sum_{p=1}^{P} \hat{\qv}_t(p)\Lc\left((\hat{\wv}_{k,t}^{p})^{\trasp}\zv_p(\xv_{k,t}),y_{k,t}\right).
\end{align*}Thus, $\{X_{k,t}: t\in [T]\}$ is a martingale difference sequence and $X_{k,t} \in [B_t, B_t+c_t]$ is bounded, where $B_t= -U_t$ is a random variable and $\Fc_{t-1}$ measurable, and $c_t = 1$. From Azuma-Hoeffding's inequality, the following bound holds for some $\delta>0$ with at least probability $1-\delta$:
\begin{align}
    \sum_{t=1}^{T}X_{k,t} \leq  \sqrt{\frac{\log{\delta^-1}}{2}T}. 
\end{align} Since this is true for any $k \in \Vc$,  we have:
\begin{equation}
    \sum_{t=1}^{T}\sum_{k=1}^{K} X_{k,t} \leq K \sqrt{\frac{\log{\delta^-1}}{2}T }.
\end{equation} This completes the proof.




\end{document}